\RequirePackage{fix-cm}
\documentclass[twocolumn]{svjour3}          
\smartqed  
\usepackage{graphicx}
\usepackage{mathptmx}      
\usepackage{algorithm}
\usepackage{algpseudocode}
\usepackage[colorlinks,linkcolor=black,anchorcolor=black,citecolor=black,urlcolor=blue]{hyperref}
\usepackage{mathtools}
\usepackage{amssymb}
\usepackage{mciteplus}
\usepackage{cite}
\usepackage{caption}
\usepackage{subcaption}

\newcommand{\Uivec}{{\mathbf{u}_i}}

 \journalname{TBA}

\begin{document}

\title{Multiscale Laplacian Learning\thanks{This work was supported in part by NIH grant GM126189, NSF Grants DMS-1721024, DMS-1761320, and IIS1900473, Michigan Economic Development Corporation, Bristol-Myers Squibb, Pfizer, and University of Kentucky Startup Fund. The authors thank the IBM TJ Watson Research Center, the COVID-19 High Performance Computing Consortium, and NVIDIA for computational assistance.
}
}


\author{Ekaterina Merkurjev         \and     Duc Duy Nguyen  \and Guo-Wei Wei }


\institute{Ekaterina Merkurjev \at
             Department of Mathematics, \\ Michigan State University, MI 48824, USA\\
              \email{merkurje@msu.edu}           
           \and
           Duc Duy Nguyen  \at
      Department of Mathematics, University of Kentucky, KY 40506, USA\\
                    \email{ducnguyen@uky.edu}       
         \and
                  Guo-Wei Wei   \at
                      Department of Mathematics, Department of Biochemistry and Molecular Biology, Department of Electrical and Computer Engineering \\
             Michigan State University, MI 48824, USA\\
                           \email{wei@math.msu.edu}       
}

\date{Received: date / Accepted: date}

\maketitle

\begin{abstract}
  Machine learning methods have greatly changed science, engineering, finance, business, and other fields. Despite the tremendous accomplishments of machine learning and deep learning methods, many challenges still remain. In particular, the performance of machine learning methods is often severely affected in case of diverse data, usually associated with smaller data sets or data related to areas of study where the size of the data sets is constrained by the complexity and/or high cost of experiments. Moreover, data with limited labeled samples is a challenge to most learning approaches. In this paper, the aforementioned challenges are addressed by integrating graph-based frameworks, multiscale structure, modified and adapted optimization procedures and semi-supervised techniques. This results in two innovative multiscale Laplacian learning (MLL) approaches for machine learning tasks, such as data classification, and for tackling diverse data, data with limited samples and smaller data sets. The first approach, called multikernel manifold learning (MML), integrates manifold learning with multikernel information and solves a regularization problem consisting of a loss function and a warped kernel regularizer using multiscale graph Laplacians. The second approach, called the multiscale MBO (MMBO) method, introduces multiscale Laplacians to a modification of the famous classical Merriman-Bence-Osher (MBO) scheme, and makes use of fast solvers for finding the approximations to the extremal eigenvectors of the graph Laplacian. We demonstrate the performance of our methods experimentally on a variety of data sets, such as biological, text and image data, and compare them favorably to existing approaches.

\end{abstract}

\section{Introduction}
 
Artificial intelligence, including machine learning, has irreversibly changed many fields including science and engineering \cite{jordan2015machine,kotsiantis2007supervised}. In fact, the combination of artificial intelligence (AI) and big data has been referred to as the “fourth industrial revolution” \cite{schwab2017fourth}. Nevertheless, machine learning tasks face several challenges.

First, while the big data challenge is well known, little attention is paid to the diverse data challenge. The success behind machine learning is that the behavior in unknown domains can be accurately estimated by quantitatively learning the pattern from sufficient training samples. However, while data sets in computer vision and image analysis often contain millions or billions of points, it is typically difficult to obtain large data sets in science \cite{jiang2020boosting}. We often deal with diverse data originating from  a relatively small data set lying in a huge space. For example, due to the complexity, ethnicity, and high cost of scientific  experiments \cite{shaikhina2017handling, shaikhina2015machine,saha2016multiple,hudson2000neural},  it is extremely difficult  to collect a relatively small set  of drug candidates of the order of 10$^6$ for a therapeutic target, while the size of the underlying chemical space of potentially pharmacologically active molecules is about 10$^{60}$ \cite{bohacek1996art}. Therefore, researchers try to cover as many components as possible with a small number of sampling points. The diversity is created by deliberately sampling a wide distribution in the huge space to understand the landscape of potential drugs. This practice is very common in scientific explorations. Similar diverse data sets exist in materials design \cite{feng2019using, zhang2018strategy}.  Overall, diverse data originated from a relatively small data set lying in a huge chemical space gives rise to a serve challenge for machine learning. Mathematically, diverse data involves disconnected submanifolds and/or nested submanifolds corresponding to multiphysics and multiscale natures of the diversity, respectively \cite{chen2020evolutionary,nguyen2020review}. The multiphysics and multiscale representations of data have been addressed by the authors' earlier work on element-specific persistent homology  \cite{ZXCang:2017a,ZXCang:2017b,ZXCang:2017c,cang2018representability}. However,  multiscale graph learning models have hardly been developed.  The proposed algorithms of this paper fill the gap, addressing the multiphysics nature of data diversity through a multiphysics data representation, such as the element-specific feature extraction developed in \cite{ZXCang:2017a, ZXCang:2017b, ZXCang:2017c, cang2018representability, nguyen2019agl}.

  Second, the success of many existing approaches for machine learning tasks, such as data classification, is dependent on a sufficient amount of labeled samples. However, obtaining enough labeled data is difficult as it is time-consuming and expensive, especially in domains where only experts can determine experimental labels; thus, labeled data is scarce. 
As a result,  the majority of the data embedded into a graph is unlabeled data, which is often much easier to obtain than labeled data but more challenging to predict. 
		Overall, one of the key limitations of most existing approaches is their reliance on large labeled sets; in particular, deep learning approaches often require massive labeled sets to learn the patterns behind the data. These challenges call for innovative strategies to revolutionize the current state-of-the-art.

Recently, algorithms involving the graph-based framework, such as those described in Section \ref{previous_work}, have recently become some of the most competitive approaches for applications ranging from image processing to the social sciences. Such methods have been successful in part due to the many advantages offered by using a graph-based approach. For example, a graph-based framework provides valuable information about the extent of similarity between elements of both labeled and unlabeled data via a weighted similarity graph and also yields information about the overall structure of the data. Moreover, in addition to handling nonlinear structure, a graph setting embeds the dimension of the features in a graph during weight computations, thus reducing the high-dimensionality of the problem. The graph framework is also able to incorporate diverse types of data, such as 3D point clouds, hyperspectral data, text, etc.

Inspired by the recent successes, we address the aforementioned challenges by integrating similarity graph-based frameworks, multiscale structure, modified and adapted optimization techniques and semi-supervised procedures, with both labeled and unlabeled data embedded into a graph. Overall, this paper formulates two multiscale Laplacian learning (MLL) approaches for machine learning tasks, such as data classification, and for dealing with diverse data, data with limited samples and smaller data sets. The first approach, the multikernel manifold learning (MML) method, introduces multiscale kernels to manifold regularization. This approach integrates new multiscale graph Laplacians into loss-function based minimization problems involving warped kernel regularizers. The second approach, the multiscale Merriman-Bence-Osher (MMBO) method, adapts and generalizes the classical Merriman-Bence-Osher (MBO) scheme \cite{merriman} to a multiscale graph Laplacian setting for learning tasks. The MMBO approach also makes use of fast solvers, such as \cite{nystrom1,nystrom2,nystrom3} and \cite{anderson}, for finding approximations of the extremal eigenvectors of the graph Laplacian. 
We validate the proposed MLL approaches using a variety of data sets. 

There are several strengths of the proposed methods:
\begin{itemize} 
 \vspace{-0.0cm}
 \item The methods address the multiscale nature of data through a multiphysics data representation, allowing them to perform well in the case of diverse data, which often occurs in, e.g., scientific applications.
     \vspace{0.1cm}
 \item The methods require less labeled training data to accurately classify a data set compared to most existing machine learning techniques, especially supervised approaches, and often in considerably smaller quantities. This is in part due to the usage of a similarity graph-based framework and the fact that the majority of the data embedded into the graph is unlabeled data.
In fact, in most cases, a good accuracy can be obtained with {\it at most} 1\%-5\% of the data elements serving as labeled data. This is an important advantage due to the scarcity of labeled data for most applications. 
  \vspace{0.1cm}
  \item Although equally applicable and successful in the case of larger data, the new methods also perform well in the case of smaller data sets, which often result in unsatisfactory performances for existing machine learning techniques, due to an often insufficient number of labeled samples and a decreased ability for machine learning-based models to learn from the observed data. 
\end{itemize}

The proposed MMBO method offers specific advantages:
\begin{itemize} 
 \vspace{-0.0cm}
\item Although it can perform just as successfully on smaller data, the MMBO algorithm is equipped with a structure which allows it to be easily adapted and designed specifically for the use of large data. In particular, in the case of large data, one can use a slight modification of the fast Nystr\"{o}m extension procedure \cite{nystrom1,nystrom2,nystrom3} to compute an approximation to the extremal eigenvectors of the multiscale graph Laplacian using a dense graph without the need to compute all the graph weights; in fact, only a small portion of the weights need to be calculated. Overall, the method uses a low-dimensional subspace spanned by only a small number of eigenfunctions. 
 \vspace{0.1cm}
\item Once the $N_e$ eigenvectors of the graph Laplacian are computed, the complexity of this algorithm is linear. Moreover, the Nystr\"{o}m extenstion procedure allows the $N_e$ eigenvectors of the graph Laplacian to be computed using only $O(N N_e)$ operations, where $N_e << N$ and $N$ is the number of data elements.
\end{itemize}

\indent The paper is organized as follows. In Section \ref{Background}, we present background, previous work and an overview of graph learning methods. In Section \ref{Methods}, we derive the proposed MML and MMBO methods and provide details on the computation of eigenvectors of the graph Laplacian for the latter method. The results from experiments are described in Section \ref{Results}, and we present a conclusion in Section \ref{conc}.

\section{Background}\label{Background}

\subsection{Previous work}
\label{previous_work}

In this section, we review recent graph-based methods for data classification and semi-supervised learning, including approaches related to convolutional neural networks, support vector machines, neural networks, label propagation, embedding methods, multi-view and multi-modal methods.

Convolutional neural networks have recently been extended to a graph-based framework for the purpose of semi-supervised learning. In particular, \cite{conv1} presents a scalable approach using graph convolutional networks by integrating a convolutional architecture motivated by a localized first-order approximation of spectral graph convolutions.   Deeper insights into the graph convolutional neural network model are discussed in \cite{conv3}. Moreover, a dual graph-based convolutional network approach is described in \cite{conv2}, while a Bayesian graph convolutional network procedure is derived in \cite{conv4}. In \cite{conv5}, a multi-scale graph convolution model is presented. In \cite{conv6}, generalizations of convolutional neural networks to signals defined on more general domains using two constructions are described; one of the generalizations is based on the spectrum of the graph Laplacian. 

    Neural networks have also been extended to a graph-based framework for the task of semi-supervised learning. For example, attention-based graph neural networks \cite{the}, graph partition neural networks \cite{gpnn}, and graph Markov neural networks \cite{qu} have been proposed.

   Moreover, support vector machines are also applied to semi-supervised learning using a graph-based framework. In \cite{svm1}, graph-based support vector machines are derived to emphasize low density regions. Also, Laplacian support vector machines (LapSVM) \cite{svm2,belkin2006manifold} and Laplacian twin support vector machines (Lap-TSVM) \cite{qi} have been formulated.

Label and measure propagation methods are discussed in, e.g., \cite{iscen}, where the authors derive a transductive label propagation method that is based on the manifold assumption. Label propagation techniques and the use of unlabeled data in classification are investigated in \cite{zhu}. Dynamic label propagation is studied in \cite{dynamic}, while semi-supervised learning with measure propagation is described in \cite{sub}.

Embedding methods are also used for semi-supervised learning.  Nonlinear embedding algorithms for use with shallow semi-supervised learning techniques, such as kernel methods, are applied to deep multi-layer architectures in \cite{weston}. Other graph embedding methods are presented in \cite{yang}.

Multi-view and multi-modal methods include \cite{nie}, which proposes a reformulation of a standard spectral learning model that can be used for multiview clustering and semi-supervised tasks. The work \cite{nie2} proposes novel multi-view learning, while \cite{gong} describes multi-modal curriculum learning. 


Other techniques for graph-based semi-supervised learning include fast anchor graph regularization \cite{wang}, a Bayesian framework for learning hyperparameters \cite{kapoor}, and random subspace dimensionality reduction.  In \cite{goldberg}, a classification method is proposed to learn from dissimilarity and similarity information on labeled and unlabeled data using a novel graph-based encoding of dissimilarity. Random graph walks are used in \cite{lin}, and sampling theory for graph signals is utilized in \cite{gadde}. In \cite{greedy}, a bivariate formulation for graph-based semi-supervised learning is shown to be equivalent to a linearly constrained max-cut problem. Lastly, reproducing kernel Hilbert spaces are used in \cite{sind}.

Various approaches involving graph-based regularization terms include regularization frameworks \cite{zhou:bousquet:lal,zhou:scholkopf}, regularization developments  \cite{chapelle:scholkopf:zien}, anchor graph regularization \cite{wang}, manifold regularization \cite{belkin2006manifold }, measure propagation \cite{sub}, approximate energy minimization \cite{boykov1}, nonlocal discrete regularization \cite{elmoataz:lezoray:bougleux}, power watershed \cite{couprie:grady:najman}, spectral matting \cite{levin:acha:lischinski}, Laplacian regularized least squares \cite{sindhwani2005beyond}, locality and similarity preserving embedding \cite{fang2014}, and clustering \cite{nie2014}. Examples for graph Laplacian regularization include label propagation \cite{zhu} and deep semi-supervised embedding \cite{weston}.

Merkurjev and co-authors have studied graph-based spectral approaches \cite{merkurjev, garcia, merkurjev_aml, merkurjev2, gloria, merkurjev_pagerank, gerhart, merkurjev_cut} using Ginzburg-Landau techniques and modifications of the MBO scheme \cite{merriman}, which is an efficient method for evolving an interface by mean curvature in a continuous setting and which can be linked to optimization problems involving the Ginzburg-Landau functional. Specifically, the MBO scheme can be derived from a Ginzburg-Landau functional minimization procedure, and can be modified and transferred to a graph setting using more general operators on graphs, as shown in Merkurjev's work on data classification  \cite{merkurjev,garcia,merkurjev2,gloria,merkurjev_aml}.

Overall, Merkurjev and co-authors have shown that multiclass data classification can be achieved using techniques from topological spaces and the Gibbs simplex \cite{garcia, merkurjev_aml}. In particular, MBO-like methods were developed for image processing applications \cite{merkurjev}, hyperspectral imaging \cite{merkurjev2, gerhart}, Cheeger and ratio cut applications \cite{merkurjev_cut}, heat kernel pagerank applications \cite{merkurjev_pagerank}, and unsupervised learning \cite{gloria}. The subject of this paper is to integrate elements of this prior work, prior work on manifold learning and novel graph-based formulations into a multiscale framework to develop new multiscale graph-based methods for machine learning tasks, such as data classification. Our methods will be able to deal with a variety of scales present in many data sets.

\subsection{Graph-based framework} \label{sec:method_setting}

The methods presented in this paper use a similarity graph framework consisting of a graph $G = (V,E)$, where $V=\{\mathbf{x}_1,\dots,\mathbf{x}_N\}$ is a set of vertices associated with the elements of the data set, and $E$ is a set of edges connecting some pairs of vertices. The edges are weighted by a weight function $w:V \times V \rightarrow \mathbb{R}$, where $w(\mathbf{x}_i,\mathbf{x}_j)$ measures the degree of similarity between $\mathbf{x}_i$ and $\mathbf{x}_j$. Larger values indicate similar elements and smaller values indicate dissimilar elements. Naturally, the embedding of data into a graph depends greatly on the edge weights. This section provides more details about graph construction, but the exact manner of weight construction for particular data sets is described in Section 4. 

The use of the graph-based framework offers many advantages. First, it provides valuable information about the extent of similarity between pairs of elements of both labeled and unlabeled data via a weighted similarity graph and also yields information about the overall structure of the data. This reduces the amount of labeled data needed for good accuracy. Moreover, a graph-based setting embeds the dimension of the features in the graph during weight computations, thus reducing the high-dimensionality of the problem. It also provides a way to handle nonlinearly separable classes and affords the flexibility to incorporate diverse types of data. In addition, in image processing, the graph setting allows one to capture texture more accurately. 

          The exact technique of computing the similarity value between two elements of data depends on the data set, but first involves feature (attribute) vector construction and a distance metric chosen specifically for the data and task at hand. For example, for hyperspectral data, one may choose the feature vector to be the vector of intensity values in its many bands and the distance measure to be the cosine distance. For 3D sensory data, one can take the feature vector to contain both geometric and color information; the weights can be calculated using a Gaussian function incorporating normal vectors, e.g., \cite{bae_merkurjev}. For text classification, popular feature extraction methods include term frequency- inverse document frequency and bag-of-words, both described in \cite{bag}. For biological data tasks, such as protein classification, persistent homology \cite{cang2018representability} can be used for feature construction. 
          
 Once the features are constructed, the weights are computed. Popular weight functions include the Zelnik-Manor and Perona function \cite{zelnik} and the Gaussian function \cite{luxberg}: 
\begin{equation}
w(\mathbf{x}_i,\mathbf{x}_j)=\exp\left(-\frac{d(\mathbf{x}_i,\mathbf{x}_j)^{2}}{\sigma^2}\right), 
\end{equation}
where $d(\mathbf{x}_i,\mathbf{x}_j)$ represents a distance between feature vectors of data elements $\mathbf{x}_i$ and $\mathbf{x}_j$, and $\sigma>0$. Using the weight function $w$, one can construct a weight matrix $\mathbf{W}$ defined as $\mathbf{W}_{ij}=w(\mathbf{x}_i,\mathbf{x}_j)$, and define the
degree of a vertex $\mathbf{x}_i\in V$ as $d(\mathbf{x}_i)= \sum_{j} w(\mathbf{x}_i,\mathbf{x}_j)$. If $\mathbf{D}$ is the diagonal matrix with elements $d(\mathbf{x}_i)$, then the graph Laplacian is defined as 
\begin{equation}
\label{Laplacian} \mathbf{L}=\mathbf{D}-\mathbf{W}. 
\end{equation}
It is sometimes beneficial to use normalized versions of the graph Laplacian, such as a symmetric graph Laplacian \cite{luxberg}.

          For some data, it is more desirable to compute the weights directly by calculating pairwise distances. In this case, the efficiency can be increased by using parallel computing or by reducing the dimension of data. Then, a graph is often made sparse using, e.g., thresholding or a nearest neighbors technique, resulting in graph where most of the edge weights are zero. Thus, the number of needed computations is reduced. Overall, a nearest neighbor graph can be computed efficiently using the $kd$-tree code of VLFeat library \cite{vlfeat}. In particular, for the nearest neighbor technique, vertices $\mathbf{x}_i$ and $\mathbf{x}_j$ are connected only if $\mathbf{x}_i$ is among the $N_n$ nearest neighbors of $\mathbf{x}_j$ or if $\mathbf{x}_j$ is among the $N_n$ nearest neighbors of $\mathbf{x}_i$. Otherwise, $w(\mathbf{x}_i,\mathbf{x}_j)$ is set to $0$. 
           
For very large data sets, one can efficiently construct an approximation to the full graph using e.g. sampling-based approaches, such as the fast Nystr\"{o}m Extension method \cite{nystrom1}.

\subsection{Semi-supervised setting}

         Despite the tremendous accomplishments of machine learning, its success depends on a sufficient amount of labeled samples. However, obtaining enough labeled data is difficult as it is time-consuming and expensive. Therefore, labeled data is scarce for most applications. 
         

However, unlabeled data is usually easier and less costly to obtain than labeled data. Therefore, it is advantageous to use a semi-supervised setting, which uses both labeled and unlabeled data to construct the graph in order to reduce the amount of labeled data needed for good accuracy. In fact, the use of unlabeled data for graph construction allows one to obtain structural information of the data. Overall, for most graph-based semi-supervised methods, the majority of data embedded into a graph is unlabeled data. This paper derives methods which use a semi-supervised setting of this kind.


\section{Methods}\label{Methods}

\subsection{Background and related graph Laplacian methods}
\subsubsection{Manifold learning}

For the derivation of the MML method, let $K$ be the number of classes, $\mathcal{L}$ be the set of labeled vertices, and $\mathcal{U}$ be the set of unlabeled vertices. We assume that $\mathcal{L}$ is drawn from the joining distribution $P$ on $V\times\mathbb{R}$, while $\mathcal{U}$ is drawn from the marginal distribution $P_V$ of $P$. We also assume that the conditional distribution $P(y|\mathbf{x})$ varies smoothly in the intrinsic geometry generated by $P_V$, where $y\in[1,K]$ and $\mathbf{x}\in V$. 





 In graph-based methods, information about labeled data and the geometric structure of the marginal distribution $P_V$ of the unlabeled samples is incorporated into the problem:
\begin{align}
f^*=\mathrm{arg}\min_{f\in \mathcal{H}_M}\frac{1}{|\mathcal{L}|}\sum_{\mathbf{x}_i\in \mathcal{L}}J(f,\mathbf{x}_i,y_i)+\gamma_A\|f\|^2_M +\gamma_I\|f\|^2_I,\label{ssl_1}
\end{align}
where the Mercer kernel $M:V\times V\rightarrow\mathbb{R}$ uniquely defines a reproducing kernel Hilbert space (RKHS) $\mathcal{H}_M$ with the corresponding norm $\|.\|_M$, $J$ is a loss function which gives rise to different types of regularization problems, $\gamma_A>0$, $\gamma_I>0$, and $\|f\|^2_I$ is an additional regularizer that reflects the intrinsic geometry of $P_V$. The solution $f^*$ to \eqref{ssl_1} can be described using the classical representer theorem \cite{representer}:
\begin{align}
f^*(\mathbf{x})=\sum_{\mathbf{x}_i\in \mathcal{L}}\alpha_i M(\mathbf{x}_i,\mathbf{x}) + \int_{\mathcal{S}} \alpha(\mathbf{z})M(\mathbf{x},\mathbf{z})dP_V(\mathbf{z}),
\end{align}
where $\mathcal{S}$ is the support of the marginal distribution $P_V$ \cite{belkin2006manifold}.

In practice, that marginal distribution is unknown. In spite of that, one could empirically estimate $\|f\|_I$  by making use of the weighted graph as discussed in Section \ref{sec:method_setting}. With the pre-defined graph Laplacian matrix $\mathbf{L}$, the manifold regularizer $\|f\|^2_I$ can be empirically estimated \cite{belkin2006manifold} as 
\begin{align}
\|f\|^2_I = \sum_{i,j=1}^{n}\left(f(\mathbf{x}_i)-f(\mathbf{x}_j)\right)^2w_{ij} = \mathbf{f}^T\mathbf{L}\mathbf{f},\label{I_norm}
\end{align}
where $\mathbf{f}=[f(\mathbf{x}_1),f(\mathbf{x}_2),\cdots,f(\mathbf{x}_n)]^T$. 


The ambient norm $\|.\|_M$ and the intrinsic norm $\|.\|_I$ in \eqref{ssl_1} can be integrated in one term under the warped kernel $\tilde{M}$ \cite{sindhwani2005beyond}. This kernel defines an alternative reproducing kernel Hilbert space $\tilde{\mathcal{H}}$ by considering a modified inner product:
\begin{align}
\langle f,g \rangle_{\tilde{\mathcal{H}}_{\tilde{M}}} = \langle f,g\rangle_{\mathcal{H}_M} + \mathbf{f}^T \mathbf{P}\mathbf{g}, \label{modified_inner_product}
\end{align}
where $\mathbf{P}$ is a positive semi-definite matrix defined on labeled and unlabeled data, $\mathbf{f}=[f(\mathbf{x}_1),f(\mathbf{x}_2),\cdots,f(\mathbf{x}_n)]^T$ and $\mathbf{g}=[g(\mathbf{x}_1),g(\mathbf{x}_2),\cdots,g(\mathbf{x}_n)]^T$. With $\langle.,.\rangle_{\tilde{\mathcal{H}}_{\tilde{M}}}$, the warped kernel $\tilde{M}$ is shown in \cite{sindhwani2005beyond} to have the following representation:
\begin{align}
\tilde{M}(\mathbf{x},\mathbf{z}) = M(\mathbf{x},\mathbf{z}) -\mathbf{M}^T_\mathbf{x}(\mathbf{I}+\mathbf{P}\mathbf{M})^{-1}\mathbf{P}\mathbf{M}_\mathbf{z},\label{warped_kernel}
\end{align}
where $\mathbf{M}=[m_{ij}]$ is the Gram matrix with $m_{ij}=M(\mathbf{x}_i,\mathbf{x}_j)$, $\mathbf{M}_\mathbf{x}$ denotes the vector $(M(\mathbf{x}_1,\mathbf{x}),M(\mathbf{x}_2,\mathbf{x}),\cdots,M(\mathbf{x}_n,\mathbf{x}))^T$, and $\mathbf{M}_\mathbf{z}$ denotes the vector $(M(\mathbf{z}_1,\mathbf{x}),M(\mathbf{z}_2,\mathbf{x}),\cdots,M(\mathbf{z}_n,\mathbf{x}))^T$.

 The regularization problem for the warped kernel $\tilde{M}$ is:
\begin{align}
f^*=\mathrm{arg}\min_{f\in \tilde{\mathcal{H}}_{\tilde{M}}}\frac{1}{|\mathcal{L}|}\sum_{\mathbf{x}_i\in \mathcal{L}}J(f,\mathbf{x}_i,y_i)+\gamma_A\|f\|^2_{\tilde{M}}.\label{ssl_2}
\end{align}
Problem \eqref{ssl_2} exploits the intrinsic geometry of $P_V$ via the data-dependent kernel $\tilde{M}$ but still makes use of the classical regularization solvers. In fact, the classical representer theorem \cite{representer} allows $f^*$ in \eqref{ssl_2} to be expressed as:
\begin{align}
f^*(\mathbf{x})=\sum_{\mathbf{x}_i\in\mathcal{L}}\alpha_i \tilde{M}(\mathbf{x},\mathbf{x}_i).
\end{align}
In practice, $\{\alpha_i\}$ are numerically determined by an appropriate optimization solver, e.g., \cite{cortes1995}.

\subsubsection{MBO reduction}
\label{MBOschemes}

For the derivation of the MMBO method, we first note that a typical learning algorithm involves finding an optimal label matrix $\mathbf{U} = (\mathbf{u}_1, \dots, \mathbf{u}_N)^T$ associated with data elements, where $\mathbf{u}_i \in \mathbb{R}^K$ represents the probability distribution over the classes for data element $\mathbf{x}_i$; the $i^{th}$ row of $\mathbf{U}$ is set to $\mathbf{u}_i$. The vector $\mathbf{u}_i$ is an element of the Gibbs simplex:
\vspace{-0.1cm}
\begin{equation}
\label{gibbs}
	\Sigma^K \coloneqq \{ (z_1, \dots, z_K) \in [0,1]^K  \vert \sum_{k=1}^K z_{k} = 1  \},
\vspace{-0.2cm}
\end{equation}
where $K$ is the number of classes. Moreover, the $k^{th}$ vertex of the simplex is given by the unit vector
$\mathbf{e}_k$. 

A general form of a graph-based problem for data classification is the minimization of $E(\mathbf{U})=R(\mathbf{U})+\textrm{Fid}(\mathbf{U})$,
where $\mathbf{U}$ is the data classification function, $R(\mathbf{U})$ is a graph-based regularization term incorporating the graph weights, and $\textrm{Fid}(\mathbf{U})$ is a term incorporating labeled points.

Not surprisingly, the choice of the regularization term has non-trivial consequences in the final accuracy. In \cite{garcia}, Garcia et al. successfully take for the regularization term a multiclass graph- based Ginzburg-Landau (GL) functional: 
\vspace{-0.1cm}
\begin{equation}
\label{eqq:MGL_SSL}
	\text{GL}(\mathbf{U}) = \frac{\epsilon}{2} \langle \mathbf{U}, \mathbf{L} \mathbf{U} \rangle + \frac{1}{2 \epsilon}\sum_{i}\left( \prod_{k=1}^K \frac{1}{4} \left \Vert \Uivec - \boldsymbol{e}_k \right \Vert_{L_1}^2\right).
	\vspace{-0.1cm}
\end{equation}
Here, $\epsilon>0$, $L$ is a normalized graph Laplacian, $K$ is the number of classes, $\langle \mathbf{U}, \mathbf{L} \mathbf{U} \rangle = \mathrm{trace}( \mathbf{U}^T \mathbf{L}\mathbf{U})$, $\Uivec$ is the $i^{th}$ row of $\mathbf{U}$, $\mathbf{\hat{u}}_i$ is a vector indicating prior class knowledge of $\mathbf{x}_i$, $\boldsymbol{e}_k$ is an indicator vector of size $K$ with one in the $k^{th}$ component and zero elsewhere, and $\mu_i$ is a parameter that takes the value of $\mu>0$ if $x_i$ is a labeled data element and zero otherwise. The variable $\boldsymbol{\hat{u}}_i = \boldsymbol{e}_k$ if $x_i$ is a labeled element of class $k$. The first (smoothing) term in \eqref{eqq:MGL_SSL} measures variations in the vector field, while the second (potential) term in \eqref{eqq:MGL_SSL} drives the system closer to the vertices of the simplex. The third (fidelity) term enables the incorporation of labeled data.

While it is possible to develop a convex splitting scheme to minimize the graph-based multiclass GL energy \cite{garcia}, a more efficient technique involves MBO reduction. Specifically, if one considers the minimization of the GL functional plus a fidelity term (consisting of a fit to elements of known class) in the continuous case, one can apply $L_2$ gradient descent resulting in a modified Allen-Cahn equation. If a time-splitting scheme is then applied, one obtains a procedure where one alternates between propagation using the heat equation with a forcing term and thresholding. In such a state, the resulting procedure has similar elements to the MBO scheme \cite{mbo}, which evolves an interface by mean curvature, in a continuous, rather than graph-based, setting. The procedure can then be transferred to a graph-based setting using \cite{merkurjev, garcia, merkurjev_aml}. Moreover, in order for the scheme to be applicable to the multiclass case, one can convert the thresholding operation to the displacement of the vector field variable towards the closest vertex in \eqref{gibbs} \cite{merkurjev, garcia, merkurjev_aml}. 


\subsection{The derivation of the multiscale setting and the proposed methods}
 
\subsubsection{Multiscale graph Laplacian operator}
  The dominance of multiscale information over the single one has been proved in various biophysic-related works, such as those involving thermal fluctuation predictions \cite{opron,xia} and binding affinity predictions \cite{nguyen}. Therefore, it is promising to explore how the multiscale approach can improve the accuracy of graph-based data classification. We examine a novel multiscale graph Laplacian in the form of
\vspace{-0.15cm}
\begin{equation}
\label{multiscale}
  \mathbf{L}_{\text{multiscale}}= \sum_{t=0}^{m} c_t \mathbf{L}_t^{p_t},
\vspace{-0.15cm}
\end{equation}
where $p_t>0$, $c_t>0$, and $L_t$ is an extended Laplacian matrix defined by $\mathbf{L}_t= \mathbf{D}_t- \mathbf{W}_t$,
where $\mathbf{D}_t$ is a degree matrix, and $\mathbf{W}_t$ is an extended adjacent graph edge matrix 
\vspace{-0.0cm}
\begin{equation}
    [\mathbf{W}_t]_{ij}= \frac{1}{\sqrt{\sigma_t}} H_t \left(\frac{||x_i-x_j||}{\sigma_t}\right) e^{-\frac{||x_i-x_j||^2}{\sigma^2_t}},
\vspace{-0.0cm}
\end{equation}
where $\sigma_t>0$ and $H_t$ is the $t^{th}$ order  Hermite polynomial. Usually, {\it only two or three multiscale Laplacian terms} in \eqref{multiscale}, i.e., $m=1$ or $m=2$, are needed to obtain a significant improvement in accuracy; by setting $m=0$ and $c_0=1$, one can restore the regular graph Laplacian discussed in \eqref{Laplacian}. In this formulation, $\sigma_t$ is automated scale filtration parameter that controls the shape of a submanifold for a data set, while $c_t$ weighs contributions from different scales.
The parameters $c_t$ and $\sigma_t$ may vary for different Hermite polynomials. 

In case of large data for which computing all the graph weights can be computationally expensive, one can use the Nystr\"{o}m extension method \cite{nystrom1,nystrom2,nystrom3} to compute approximations to the few smallest eigenvalues and corresponding eigenvectors of the multiscale graph Laplacian while calculating only a small fraction of the graph weights. We will modify the Nystr\"{o}m procedure to incorporate the new multiscale graph Laplacian $\mathbf{L}_{\text{multiscale}}$. In this case, the weights in the procedure are computed using 
\vspace{-0.05cm}
\begin{equation}
\mathbf{W}_{\text{multiscale}}= \sum_{t=0}^{m} c_t \mathbf{W}_t^{p_t},
\vspace{-0.15cm}
\end{equation}
where, in most cases, $m=1$ or $m=2$ is enough to obtain a significant accuracy improvement.

When the number of data elements is not too large, one can compute the eigenvectors via the Rayleigh-Chebyshev method \cite{anderson} or the Shifted Block Lanczos algorithm \cite{lanczos}.

\subsubsection{Multikernel manifold learning (MML) scheme}

 In multikernel manifold learning (MML), the multiscale Laplacian matrices proposed in (\ref{multiscale}) is employed to form $N_n$-nearest neighbors subgraphs. By setting $\mathbf{P}=\frac{\gamma_I}{\gamma_A}\mathbf{W}_{\text{multiscale}}$ in (\ref{warped_kernel}), we attain an MML scheme enabling the reconstruction of the regularization problem presented in (\ref{ssl_1}).
Even with the integration of multiscale Laplacian operator into the data kernel, the manifold learning algorithms still retains its classical representation as presented in (\ref{ssl_2}). One, therefore, could utilize traditional solvers to derive the multiscale manifold learning's optimizer \cite{sindhwani2005beyond}. The MML procedure is summarized as Algorithm \ref{algo:MML}.

\begin{algorithm}[H]
\caption{MML Algorithm (multiscale)} \label{algo:MML}
\begin{algorithmic}[H]
\vspace{0.075cm}
\Require  labeled data $\mathcal{L}=\{(\mathbf{x}_i, y_i)\}_i$, where $y_i$ is the label of $\mathbf{x}_i$, unlabeled data $\mathcal{U}=\{\mathbf{x}_j\}_j$, $N_n$ (\# of nearest neighbors), $m+1$ (\# of scales), where $2$ or $3$ scales is usually sufficient, $\{c_t\}_{t=0}^m$ (Laplacian matrix coefficients), $\{p_t\}_{t=0}^m$ (matrix powers), $\{\sigma_t\}_{t=0}^m$ (kernel scales), and $\gamma_I>0$, $\gamma_A>0$ (scalars).
\vspace{0.1cm}
\Ensure  Estimated optimizer $f^*$, where $f^*(\mathbf{x})$ is the prediction for $\mathbf{x}$.
\vspace{0.1cm}
\State 1: Construct $m+1$ multiscale subgraphs with $N_n$ nearest neighbors with weights $[\mathbf{W}_t]_{ij}$ for $t=0,...,m$, where it is usually sufficient to use $m=1$ or $m=2$, i.e. two or three scales.
\vspace{0.1cm}
\State 2: Select the kernel $M(\mathbf{x},\mathbf{x}_i)$, e.g., radial basis function kernel or a Gaussian kernel.
\vspace{0.1cm}
\State 3: Compute the Gram matrix $\mathbf{M}=[m_{ij}]$ with $m_{ij}=M(\mathbf{x}_i,\mathbf{x}_j)$. 
\vspace{0.1cm}
\State 4: Compute the multiscale Laplacian $\mathbf{L}_{\text{multiscale}}$ using \eqref{multiscale} and $\{c_t\}_{t=1}^m$, $\{p_t\}_{t=0}^m$ and $\{\sigma_t\}_{t=0}^m$.
\vspace{0.1cm}
\State 5: Formulate the warped kernel $\tilde{M}(\mathbf{x},\mathbf{x}_i)$ using \eqref{warped_kernel} and $\mathbf{P}=\mathbf{L}_{\text{multiscale}}$.
\vspace{0.1cm}
\State 6: Solve for optimizer of (\ref{ssl_2}) using an SVM quadratic programing solver for soft margin loss, e.g., \cite{cortes1995}.
\vspace{0.1cm}
\end{algorithmic}
\end{algorithm}

\subsubsection{Multiscale MBO (MMBO) scheme}
\label{steps_MBO}

Our proposed MMBO scheme uses a semi-implicit approach where the multiscale Laplacian term is computed implicitly due to the stiffness of the operator which is caused by a wide range of its eigenvalues. An implicit term is needed since an explicit scheme requires all scales of eigenvalues to be resolved numerically. 

To derive the MMBO scheme, let $\mathbf{U}$ represent a matrix where each row is a probability distribution of each data element over the classes and let $\Uivec$ represent the $i^{th}$ row of $\mathbf{U}$. In addition, let $N$ be the number of data set elements, $K$ be the number of classes, $\text{dt}>0$, and $\boldsymbol{\mu}$ be a vector which takes a value $\mu$ in the $i^{th}$ place if $\mathbf{x}_i$ is a labeled element and $0$ otherwise. Moreover, let $\mathbf{U_\text{labeled}}$ be the following matrix: for rows corresponding to labeled points, the entry corresponding to the class of the labeled point is set to 1.  All other entries of the matrix are set to 0. Lastly, let $\boldsymbol{\mu} \cdot (\mathbf{U}-\mathbf{U_\text{labeled}})$ indicate row-wise multiplication by a scalar. 

As described in Section \ref{MBOschemes}, if one considers the minimization of a GL functional plus a fit to elements of known class in the continuous case, an $L_2$ gradient descent results in a modified Allen-Cahn equation. If a time-splitting scheme is then applied, one obtains a procedure where one alternates between propagation using the heat equation with a forcing term and thresholding. The scheme can then be transferred to a graph-based setting and the Laplace operator can be replaced by a graph-based multiscale Laplacian. The thresholding can be changed to the displacement of the variable towards the closest vertex in \eqref{gibbs}. A projection to the simplex is then necessary before the displacement step.  


Our proposed MMBO procedure thus consists of the following procedure. Starting with an initial guess for $\mathbf{U}$, obtain the next iterate of $\mathbf{U}$ via the following three steps:
\begin{enumerate}
\item Multiscale heat equation with a forcing term: \hspace{0.1cm} $ \mathbf{U}^{n+\frac{1}{2}} =\mathbf{U}^{n} -\text{dt} \{\mathbf{L_{\text{multiscale}}} \mathbf{U}^{n+\frac{1}{2}}+\boldsymbol{\mu} \cdot (\mathbf{U}^{n}-\mathbf{U_\text{labeled}})\}$, \\ where $\boldsymbol{\mu}$ is a vector which takes a value $\mu$ in the $i^{th}$ place if $\mathbf{x}_i$ is a labeled element and $0$ otherwise, and $\boldsymbol{\mu} \cdot (\mathbf{U}^{n}-\mathbf{U_\text{labeled}})$ indicates row-wise multiplication by a scalar. 
\item Projection to simplex: Each row of $\mathbf{U}^{n+\frac{1}{2}}$ is projected onto the simplex using \cite{chen}.
\item Displacement: $ \Uivec^{n+1}= \boldsymbol{e}_k$,
where $\Uivec^{n+1}$ is the $i^{th}$ row of $\mathbf{U}^{n+1}$, and $\boldsymbol{e}_k$ is the indicator vector (with a value of 1 in the $k^{th}$ place and 0 elsewhere) associated with the vertex in the simplex closest to the $i^{th}$ row of the projected $\mathbf{U}^{n+\frac{1}{2}}$ from step 2.
\end{enumerate}
\vspace{0.05cm}
This implicit scheme allows the evolution of $\mathbf{U}$ to be numerically stable regardless of the time step $dt$, in spite of the ``stiffness'' of the differential equations which
could otherwise force $dt$ to be impractically small.

One can compute $\mathbf{U}^{n+\frac{1}{2}}$ very efficiently using spectral techniques and projections onto a low-dimensional subspace spanned by a small number of eigenfunctions in the following manner, where $\mathbf{I}$ is the identity: 
\begin{equation}
\label{update}
\mathbf{U}^{n+\frac{1}{2}} = \mathbf{X}_{\text{multiscale}}\left(\mathbf{I} + \text{dt}\,\boldsymbol{\Lambda}_{\text{multiscale}}\right)^{-1} \mathbf{X}_{\text{multiscale}}^T \mathbf{U_\text{update}},
\end{equation}
where $\mathbf{U_\text{update}}= \mathbf{U}^n -\text{dt} \, \boldsymbol{\mu} \cdot (\mathbf{U}^n-\mathbf{U_\text{labeled}})$, $\mathbf{X}_{\text{multiscale}}$ is an $N \times N_e$ truncated matrix retaining only $N_e << N$ smallest eigenvectors of the multiscale graph Laplacian $\mathbf{L}_{\text{multiscale}}$, and $\boldsymbol{\Lambda}_{\text{multiscale}}$ is a $N_e \times N_e$ diagonal matrix retaining the smallest eigenvalues of $\mathbf{L}_{\text{multiscale}}$ along the diagonal.


The proposed MMBO procedure is detailed as Algorithm \ref{algo:iter_MBO}. It is important to note that in the MMBO method, the graph weights are only used to compute the few eigenvectors and eigenvalues of the multiscale graph Laplacian, and the multiscale MMBO procedure themselves do not involve graph weights. This crucial property allows one to use the Nystr\"{o}m extension procedure \cite{nystrom1,nystrom2,nystrom3} to approximate the extremal eigenvectors of the Laplacian by only computing a small portion of the graph weights; this enables one to apply the multiscale models very efficiently on large data. 

For initialization, the rows of $\mathbf{U}$ corresponding to labeled points are
set to the vertices of the simplex corresponding to the known labels,
while the rows of $\mathbf{U}$ corresponding to the rest of the points initially represent random probability distributions over the classes.

The energy minimization proceeds until a steady state condition is reached. The final classes are obtained by assigning class $k$ to node $i$ if $\Uivec$ is closest to vertex $\boldsymbol{e}_k$ on the Gibbs simplex. Consequently, the calculation is stopped when, for a positive constant $\eta>0$, 
\begin{equation}
\frac{ \max\limits_i \Vert \Uivec^{n+1} - \Uivec^{n} \Vert^{2}}{ \max\limits_i
\Vert \Uivec^{n+1} \Vert^{2}} < \eta.
\label{eq:stop}
\end{equation}

In regards to computational complexity, in practice, once the $N_e$ eigenvectors of the graph Laplacian are computed, the complexity of the MMBO scheme is linear in the number of data elements $N$. In particular, let $K$ be the number of classes and $m+1$ be the number of terms in the multiscale Laplacian \eqref{multiscale}. Usually, $m=1$ or $m=2$ is enough to obtain a good accuracy. Then, one needs $O(N K N_e )$ operations for the multiscale heat equation with a forcing term, $O(N K \log K)$ operations for the projection to the simplex and $O(N K)$ operations for the displacement step. Moreover, \cite{nystrom1,nystrom2,nystrom3} allows one to compute the $N_e$ eigenvectors of the multiscale graph Laplacian using $O(N N_e m)$ operations. Since $N_e << N$ and $K << N$, in practice, the complexity of this method is linear. 

\subsection{Computation of eigenvalues and eigenvectors of the multiscale graph Laplacian}

 The MMBO method requires one to compute a few of the smallest eigenvalues and the corresponding eigenvectors of the multiscale graph Laplacian to form $\mathbf{X}_{\text{multiscale}}$. We examine and use three techniques for this task. Nystr\"{o}m extension \cite{nystrom1,nystrom2,nystrom3} is the preferred method for very large data. 

\begin{algorithm}[H]
\caption{MMBO Algorithm (multiscale)} \label{algo:iter_MBO}
\begin{algorithmic}[H]
\vspace{0.075cm}
\Require labeled data $\mathcal{L}=\{(\mathbf{x}_i, y_i)\}_i$, where $y_i$ is the label of $\mathbf{x}_i$, unlabeled data $\mathcal{U}=\{\mathbf{x}_j\}_j$, $N_n$ (\# of nearest neighbors), $m+1$ (\# of scales), $\{c_t\}_{t=0}^m$ (Laplacian matrix coefficients), $\{p_t\}_{t=0}^m$ (matrix powers), $\{\sigma_t\}_{t=0}^m$ (kernel scales), $\text{dt}>0$, $N$ (\# of data set elements), $N_e << N$ (\# of eigenvectors to be computed), $N_t$ (maximum \# of iterations), $\boldsymbol{\mu}$ (an $N \times 1$ vector which takes a value $\mu$ in the $i^{th}$ place if $x_i$ is a labeled element and $0$ otherwise). 
\vspace{0.1cm}
\Ensure $\mathrm{out} = \mathbf{U^{end}}$; the $i^{th}$ row of $\mathbf{U^{end}}$ is a probability distribution of data element $\mathbf{x}_i$ over the classes. 
\vspace{0.1cm}
\State 1. For larger data, go to Step 4. For smaller data, go to Step 2.
\vspace{0.1cm}
\State 2: Construct $m+1$ multiscale subgraphs with $N_n$ nearest neighbors with weights $[\mathbf{W}_t]_{ij}$ for $t=0,...,m$, where it is usually sufficient to use $m=1$ or $m=2$, i.e. two or three scales.
\vspace{0.1cm}
\State 3: Compute the multiscale Laplacian $\mathbf{L}_{\text{multiscale}}$ using \eqref{multiscale} and $\{c_t\}_{t=0}^m$, $\{p_t\}_{t=0}^m$ and $\{\sigma_t\}_{t=0}^m$.
\vspace{0.1cm}
\State 4: Compute $\mathbf{U_\text{labeled}}, \boldsymbol{\Lambda_\text{multiscale}}$ and $\mathbf{X_\text{multiscale}}$ as described in Section \ref{steps_MBO} and using $N_e << N$. For smaller data, use methods such as \cite{anderson}. For larger data, use Nystr\"{o}m extension \cite{nystrom1,nystrom2,nystrom3}.
\vspace{0.1cm}
\State 5: Complete the following steps: starting with $n=1$. 
\vspace{0.1cm}
\For{$i = 1 \to N$}
\State $\mathbf{U}_{i k}^{~0} \leftarrow rand((0,1)), \Uivec^0 \leftarrow projectToSimplex(\Uivec^0)$ using \cite{chen}, where $i^{th}$ row of $\mathbf{U}^0$.
\State $\mathrm{If~} \mu_i > 0, ~ \mathbf{U}_{i k}^{~0} \leftarrow \mathbf{{U_{\text{labeled}}}}_{i k}$
\EndFor
\State $\mathbf{B} \leftarrow \left(\mathbf{I} +{\text{dt}}\boldsymbol{\Lambda_\text{multiscale}}\right)^{-1}\mathbf{X_\text{multiscale}}^T$
\While{$\mathrm{Stop~criterion~not~satisfied \hspace{0.2cm} or \hspace{0.2cm}}$$n > N_t$}
\State $\mathbf{C} \leftarrow  \mathbf{U}^n - {\text{dt}} \, \boldsymbol{\mu}(\mathbf{U}^n-\mathbf{U_\text{labeled}}) $
\State $\mathbf{A} \leftarrow \mathbf{B}\mathbf{C}$
\State $\mathbf{U}^{n+1} \leftarrow \mathbf{X_\text{multiscale}}\mathbf{A}$
\For{$i =1 \to N$}
\State $\Uivec^{n+1} \leftarrow projectToSimplex(\Uivec^{n+1})$ using \cite{chen}
\State $\Uivec^{n+1} \leftarrow \boldsymbol{e}_k$, where $k$ is closest simplex vertex to $\Uivec^{n+1}$
  \EndFor
  \State The matrix $\mathbf{U}^{n+1}$ is such that its $i^{th}$ row is $\Uivec^{n+1}$.
  \State $ n \leftarrow n + 1$
 \EndWhile
 \vspace{0.1cm}
\end{algorithmic}
\end{algorithm}

\subsubsection{Nystr\"{o}m extension for fully connected graphs} 
\label{nystrom}

Nystr\"{o}m extension \cite{nystrom1,nystrom2,nystrom3} is a matrix completion method, and it performs faster than many other techniques because it computes approximations to eigenvectors and eigenvalues using {\it much smaller} submatrices of the original matrix. 

         Note that if $\lambda$ is an eigenvalue of $\hat{\mathbf{W}}= \mathbf{D}^{-\frac{1}{2}}\mathbf{W}\mathbf{D}^{-\frac{1}{2}}$, then $1- \lambda$ is an eigenvalue of the symmetric Laplacian $\mathbf{L_{s}}=  I-\mathbf{D}^{-\frac{1}{2}}\mathbf{W}\mathbf{D}^{-\frac{1}{2}}$, and the two matrices have the same eigenvectors. Thus, one can use Nystr\"{o}m extension to calculate approximations to the eigenvectors of $\hat{\mathbf{W}}$ and thus of $\mathbf{L_{s}}$.

 Now, consider the problem of approximating the extremal $N_e$ eigenvalues and eigenvectors of a full graph $\hat{\mathbf{W}}$ and let $\hat{w}(x_i,x_j)=\hat{\mathbf{W}}_{ij}$. Nystr\"{o}m extension \cite{nystrom1,nystrom2,nystrom3} approximates the eigenvalue equation using a quadrature rule and $N_e << N$ randomly chosen interpolation points from $V$, which represents data elements. Denote the set of $N_e$ randomly chosen points by $X= \{x_{i}\}^{N_e}_{i=1}$ and its complement by $Y$. Partitioning $V$ into $V= X\cup Y$ and letting $\phi_{k}(x)$ be the the $k^{th}$ eigenvector of $W$ and $\lambda_{k}$ be its associated eigenvalue, we obtain: 
\begin{equation}
        \sum_{x_{j}\in X} \hat{w}(y_{i},x_{j}) \phi_{k}(x_{j})= \lambda_{k} \phi_{k}(y_{i})   \hspace{0.2cm}   \forall  y_{i} \in Y, \hspace{0.1cm} \forall k \in {1,..., N_e}.
\end{equation}
This system cannot be solved directly since the eigenvectors are unknown; thus, the $N_e$ eigenvectors of $\hat{\mathbf{W}}$ are approximated using much smaller submatrices of $\hat{\mathbf{W}}$. 

  The efficiency of Nystr\"{o}m extension lies with the following fact: when computing the $N_e$ eigenvalues and eigenvectors of an $N \times N$ matrix, where $N$ is large, the algorithm approximates them using much smaller matrices, the largest of which has dimension $N \times N_e$, where $N_e << N$. 
In particular, when the method is applied to $\mathbf{W}$ or $\hat{\mathbf{W}}$, {\it only a small portion of the weight matrix $\mathbf{W}$ or $\hat{\mathbf{W}}$ needs to be computed}. In our experience, $N_e=100$ or $N_e=200$ were good choices. 
 
If the number of scales is $m+1$, the complexity of the Nystr\"{o}m procedure is  $O(N N_e (m+1))$, which is linear in $N$. 

\subsubsection{ Rayleigh-Chebyshev method} \label{sec:rc} The Rayleigh-Chebyshev method \cite{anderson} is a fast algorithm for finding a small subset of eigenvalues and eigenvectors of sparse symmetric matrices, such as a symmetric graph Laplacian which can be made sparse using techniques such as $N_n$-nearest neighbors. The method is a modification of an inverse subspace iteration procedure and uses adaptively determined Chebyshev polynomials. 

\subsubsection{A shifted block Lanczos algorithm} A shifted block Lanczos algorithm \cite{lanczos}, as well as other variations of the Lanczos method \cite{old} that is an adaptation of power methods, are efficient techniques for solving sparse symmetric eigenproblems and for finding a few of the extremal eigenvalues. They can be used to find a subset of the eigenvalues and eigenvectors of the symmetric graph Laplacian which can be made sparse using $N_n$-nearest neighbors. 

\section{Results and discussion}\label{Results}
\subsection{Data sets}
In this work, we validate the proposed MML and MMBO methods against three common data sets:
\begin{itemize}
\item G50C is an artificial data set inspired by \cite{grandvalet2005semi} and generated from two normal covariance Gaussian distributions. This data set has 550 data points located in $\mathbb{R}^{50}$ and two labels $\{-1,+1\}$.
\item USPST data set includes images of handwritten digits taken from the USPS test data set. This data has 2007 images to be classified into ten labels corresponding to ten numbers from 0 to 9.

\item Mac-Win data set categorizes documents, taken from 20-Newsgroups data, into 2 classes: mac or windows\cite{szummer2002partially}. This set has 1946 elements and each element is represented by a vector in $\mathbb{R}^{7511}$.

\item WebKB data set is taken from the web documents of the CS department of four universities and has been used extensively. It has 1051 data samples and two labels: course and non-course. There are two ways to describe each web document: the textual content of the webpage (called page representation), and the anchor text on hyperlinks pointing from other webpages to the current one. The data points with page representation are in $\mathbb{R}^{3000}$, while the ones with link representation belong to $\mathbb{R}^{1840}$. When we combine two different kinds of representations, we achieve the data points in $\mathbb{R}^{4840}$.

\item $\alpha,\beta$-protein data set consists of three different protein domains, namely alpha proteins, beta proteins, and mixed alpha and beta proteins, classified based on protein secondary structures \cite{ZXCang:2015}. This data has 900 biomolecules, and each family has 300 instances.

\end{itemize}

The details of the data sets are outlined in Table \ref{tab:datasets}.

\begin{table*}[!ht]
	\centering
	{
		\caption{Data sets used in the experiments.}
		\label{tab:datasets}
		\vspace{0.05cm}
		\begin{tabular}{lcccc}
			\hline
			Data set & No. of classes & Sample dim. & No. of data elements & No. of labeled data\\
			\hline
			G50C	&	2			 &	50         &	550		 &		50\\
			USPST   & 20			 &  256 	   &	2007	 &		50\\
			Mac-Win &	2			 & 	7511 	   &	1946 	 &		50\\
			WebKB (page) & 2         &  3000       &    1051     &      12\\
			WebKB (link) & 2         &  1840       &    1051     &      12\\
			WebKB (page+link) & 2    &  4840       &    1051     &      12\\       
			$\alpha,\beta$-protein & 3 & 50        &    900      &  720\\
			\hline 
		\end{tabular}
	}
\end{table*}

\subsection{Hyperparameters selection}
In the MMBO setting, for each data point, we do not compute the complete graph but instead construct a $N_n$-nearest neighbor graph for the calculation efficiency. The parameter $l$ is one of the hyperparameters and is selected on a case by case basis. Moreover, as discussed in Section \ref{sec:method_setting}, the weight function used is the Gaussian kernel $w(\mathbf{x}_i,\mathbf{x}_j)=\exp(-d(\mathbf{x}_i,\mathbf{x}_j)^2/\sigma^2)$. Here, the scalar $\sigma$ is optimized so that it perfectly fits the labeled set information. In the multiscale approach, each kernel is assigned different $\sigma$ values depending on the outcome of hyperparameter selection. Overall, due to the random initialization of the non-labeled points, we use the same random seed for all the experiments in this work for reproducible purposes. 

The Nystr\"{o}m extension method \cite{nystrom1, nystrom2, nystrom3} allows for fast computations even in case of larger data since this approach approximates the eigenvalues and eigenvectors of the origin matrix using much smaller matrices randomly selected from the bigger ones. Thus, only a small portion of the graph weights need to be computed. However, in case of smaller data, it is often more advantageous to use methods such as \cite{anderson} which can directly compute the eigenvalues and eigenvectors. Therefore, to obtain optimal results, we employ the Rayleigh-Chebyshev procedure \cite{anderson} (see Section \ref{sec:rc}) for our experiments. This method is well-known for efficiently calculating the smallest eigenvectors of a sparse symmetric matrix. The hyperparameters of the MMBO models are the number of leading eigenvalues ($N_e$), the time step for solving heat equation ($dt$), the constraint constant on fidelity term ($\mu$), and the number of iterations ($N_t$).

The hyperparameter selection for MML model is carried out in a similar fashion as that of the MMBO algorithm. The tunning parameters are: the number of nearest neighbors ($N_n$), the scaler factor ($\sigma$), the penalty coefficient ($\gamma_A$), the manifold regularizer constraint ($\gamma_I$), and the Laplacian degree ($p$). The optimizer is solved using the primal SVM solver \cite{svm2}. The optimal hyperparameters of the proposed methods are documented in the Supporting Information.

\subsection{Performance and discussion}

\subsubsection{Non-biological data sets}
The non-biological data sets we used for our experiments are the G50C, USPST, Mac-Win, and WebKB data sets. In the experiments involving these data sets, we utilize the original representations without carrying out any feature generation procedures. In addition, following the previous work \cite{svm1,sind}, we only consider accuracy as the main evaluation metric
for the G50C, USPST, and Mac-Win data sets, and compute the Precision/Recall Breakeven Point (PRBEP) for the WebKB data set due to its imbalanced labeling. 

 In all cases, the results of the proposed MML and MMBO methods show promising improvements from non-multiscale frameworks. Specifically, the best performances of the algorithms are achieved with three kernels.  In particular, there is a significant accuracy improvement from single kernel to two kernel architectures on the USPST data (from 86.11\% to 90.57\% for the MML model, and from 86.55\% to 88.65\% for the MMBO model) and Mac-Win data (from 89.98\% to 90.01\% for the MML model, and from 92.06\% to 93.49\% for the MMBO model). The improvement from single kernel to multi-kernel learning is less for the G50C and WebKB data, but that is to be expected since G50C is a small data set consisting of 550 samples. Furthermore, it is an artificial data set drawn from two unit covariance normal distributions. As a result, a single kernel is enough to capture the crucial structure of data. Moreover, the WebKB data poses a challenge for multiscale learning due to its imbalanced data.

In almost all experiments, the proposed models obtain the best results. In particular, in most experiments, the proposed MMBO model obtains the best results, all with three-kernel learning, while the proposed MML model obtains the best result for USPST. In particular, for G50C, the MMBO method achieves the best accuracy (95.06\%), but the MML method is still comparable with its accuracy being 94.56\%. Moreover, the superior performance of our proposed algorithms over the state-of-the-art models is also displayed in the case of the more complex USPST data, a set of handwritten digit images with 1440 samples. While the proposed MML algorithm obtains the best accuracy at 90.57\%, the MMBO method with three-kernel information still obtains a good accuracy of 88.73\%. The other published approaches, such as LapRLS, obtain lower accuracies. For Mac-Win, our multi-scale models perform slightly lower than $\nabla$TSVM (94.3\%) \cite{svm1} and LDS (94.9\%) \cite{svm1}; the fact that there are only 1966 samples but the dimension of each sample is very high, i.e., 7511, might indicate noisy information which can reduce the performances of graph-based kernel models. For WebKB, our proposed methods perform extremely well. WebKB is the last data set in this category and has three different feature representations, namely, link, page and page+link.  The overall performance of our proposed models is very encouraging. We see that using only one kernel already produces great results, with a little improvement in using multiple kernels. 
The best model is the MMBO method with 3 kernels which obtains a PRBEP at 96.22\%, 97.93\%, and 98.87\% for the link, page, and page+link experiments, respectively. The MML method obtains the next best result with a PRBEP of 95.75\%, 95.81\%, and 95.84\% for the link, page, and page+link experiments, respectively. After the proposed MMBO and MML methods, the next best result is obtained by LapSVM: 94.3\%, 93.4\%, and 94.9\%.  
 
\begin{figure*}
  \vspace{0.5cm}
 \hspace{1.5cm}
 \begin{subfigure}{.4\textwidth}
  \centering
\includegraphics[width=.75\linewidth]{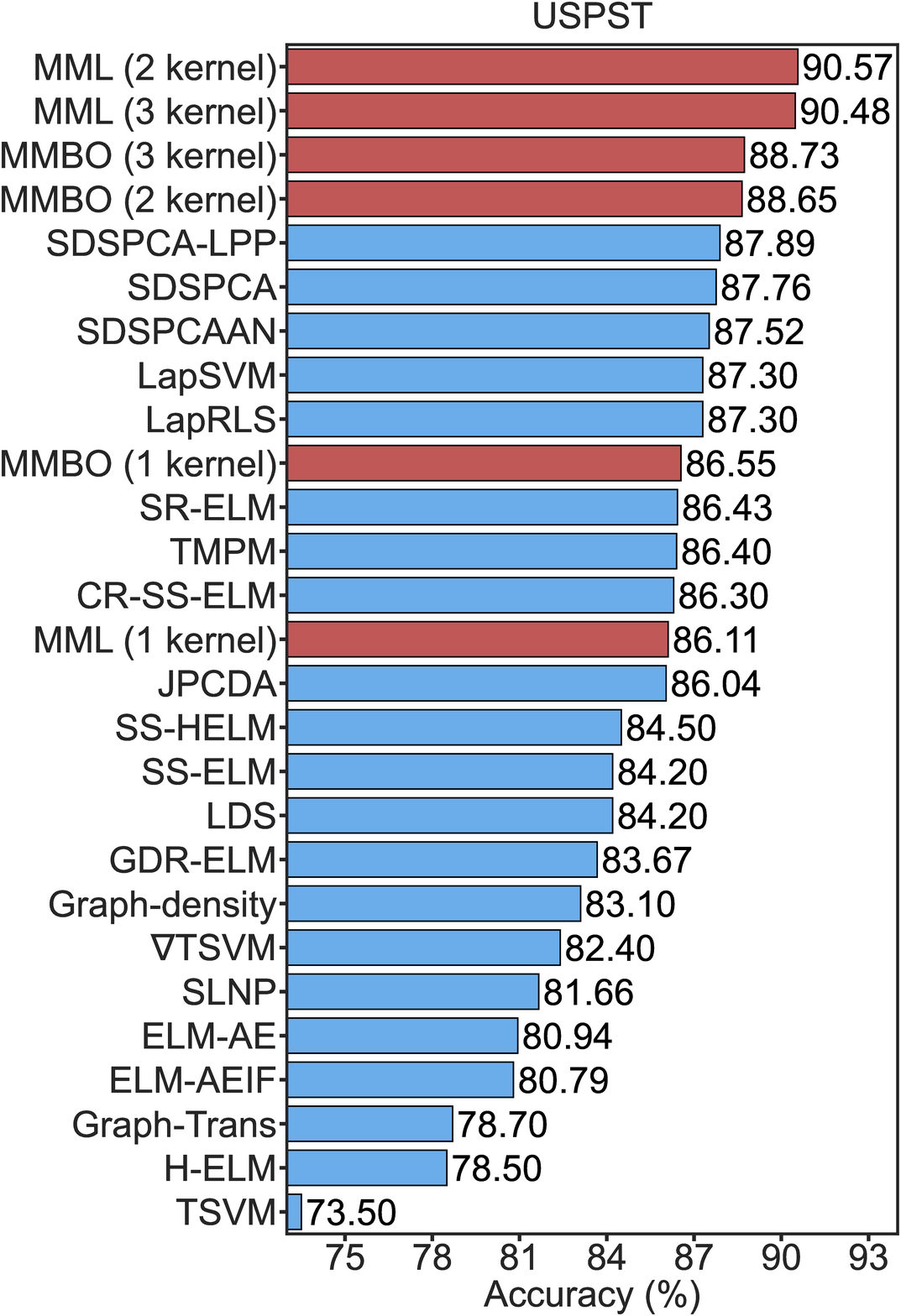}
  \label{fig:sub1}
\end{subfigure}%
\hspace{0cm}
\begin{subfigure}{.4\textwidth}
  \centering
  \includegraphics[width=.75\linewidth]{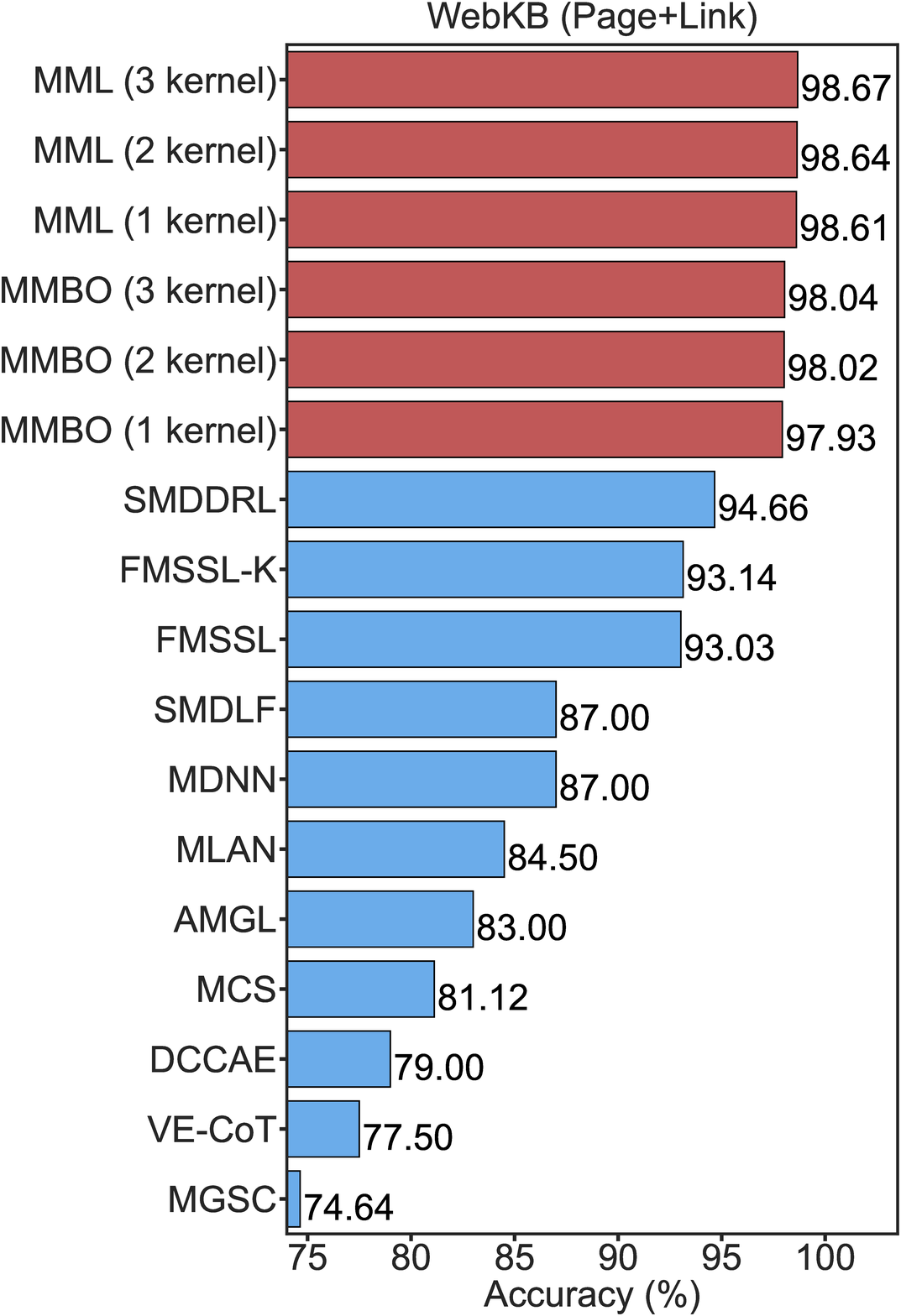}
  \label{fig:sub2}
\end{subfigure}

\vspace{0.8cm}
 \hspace{1.5cm}
\begin{subfigure}{.4\textwidth}
  \centering
  \includegraphics[width=.75\linewidth]{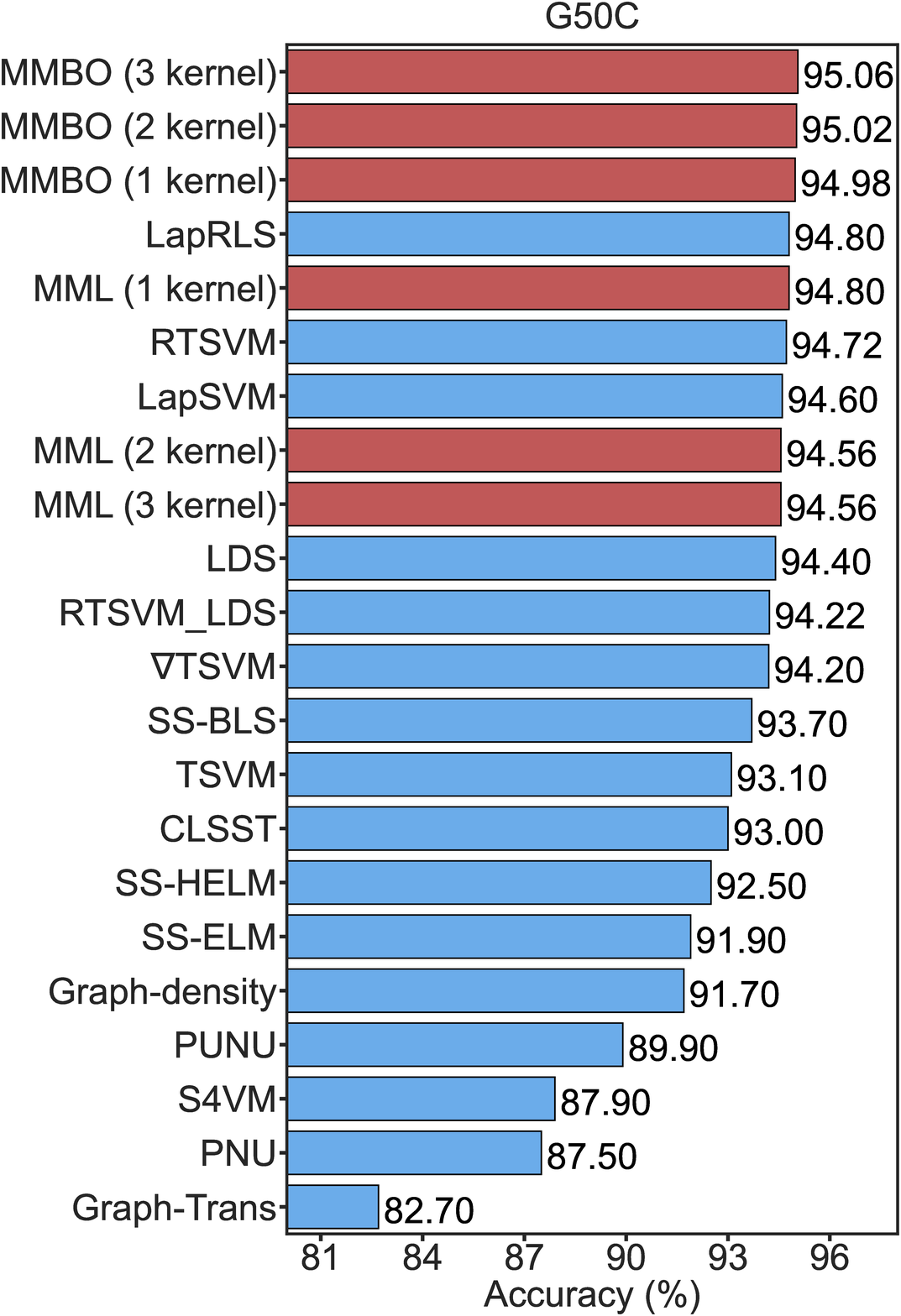}
  \label{fig:sub2}
\end{subfigure}%
\hspace{0cm}
 \begin{subfigure}{.4\textwidth}
  \centering
\includegraphics[width=.75\linewidth]{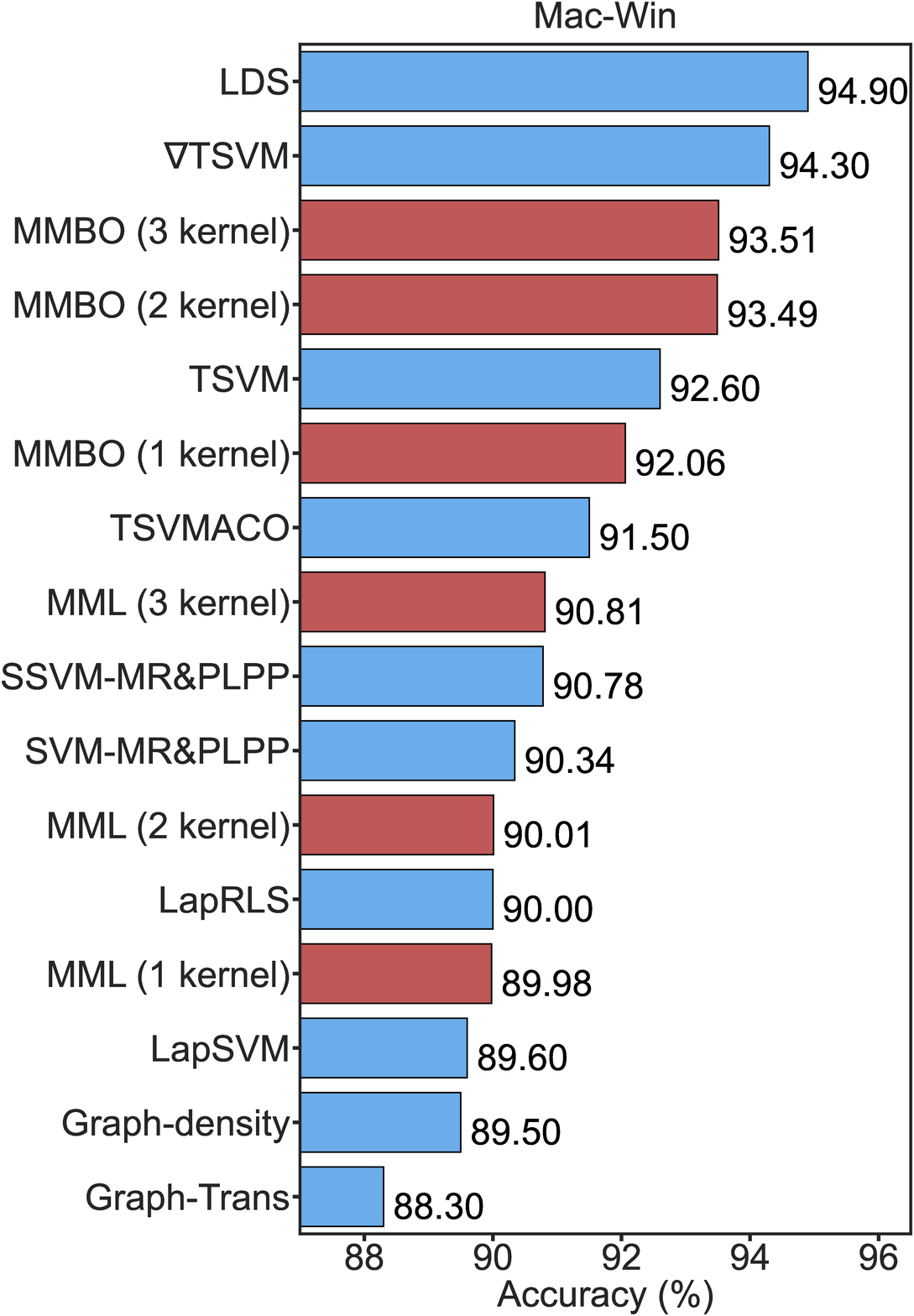}
  \label{fig:sub1}
\end{subfigure}

\vspace{0.8cm}
 \hspace{0.3cm}
\begin{subfigure}{.4\textwidth}
  \centering
  \includegraphics[width=.7\linewidth]{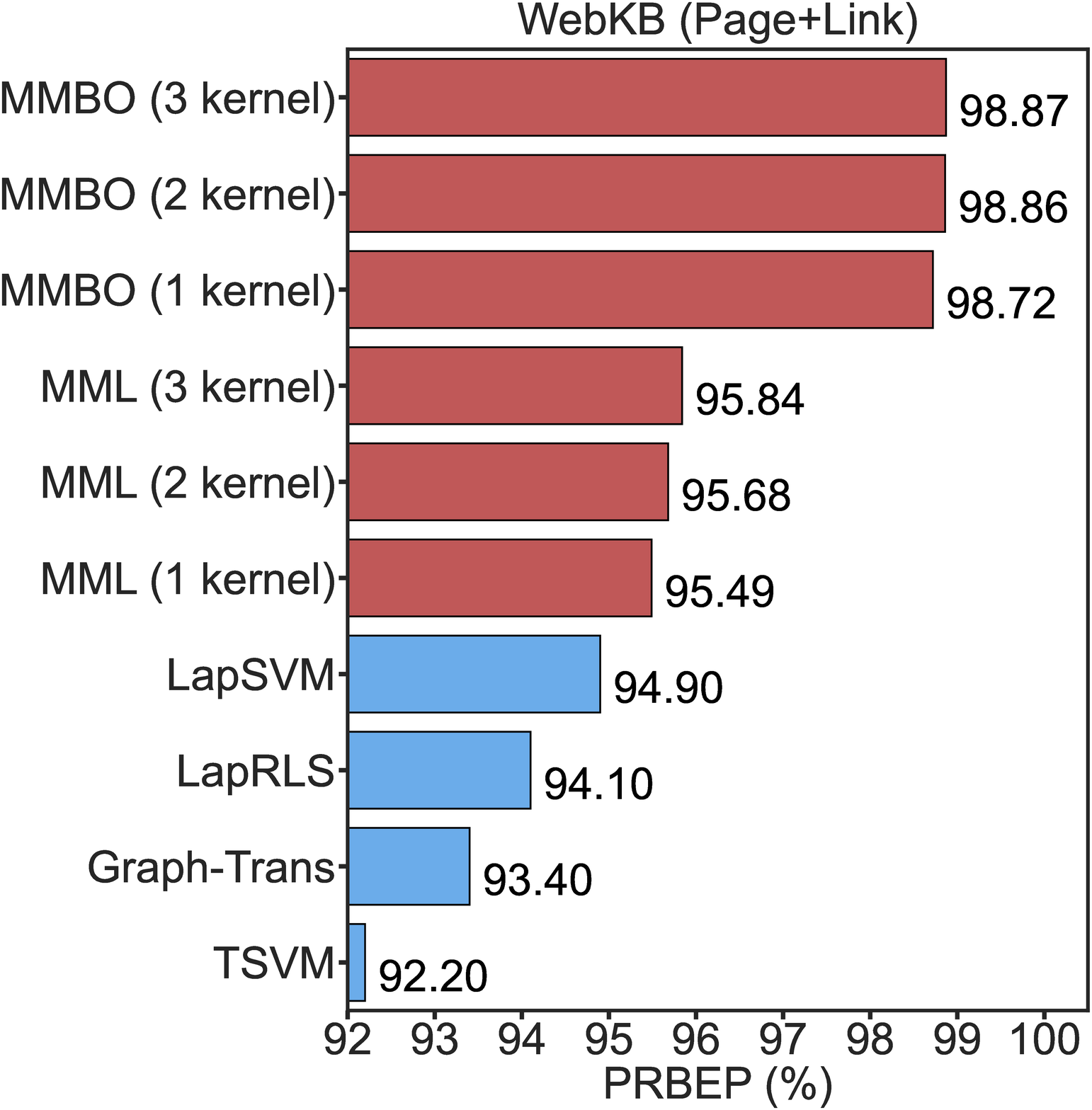}
  \label{fig:sub2}
\end{subfigure}%
\hspace{-2cm}
\begin{subfigure}{.4\textwidth}
  \centering
  \includegraphics[width=.7\linewidth]{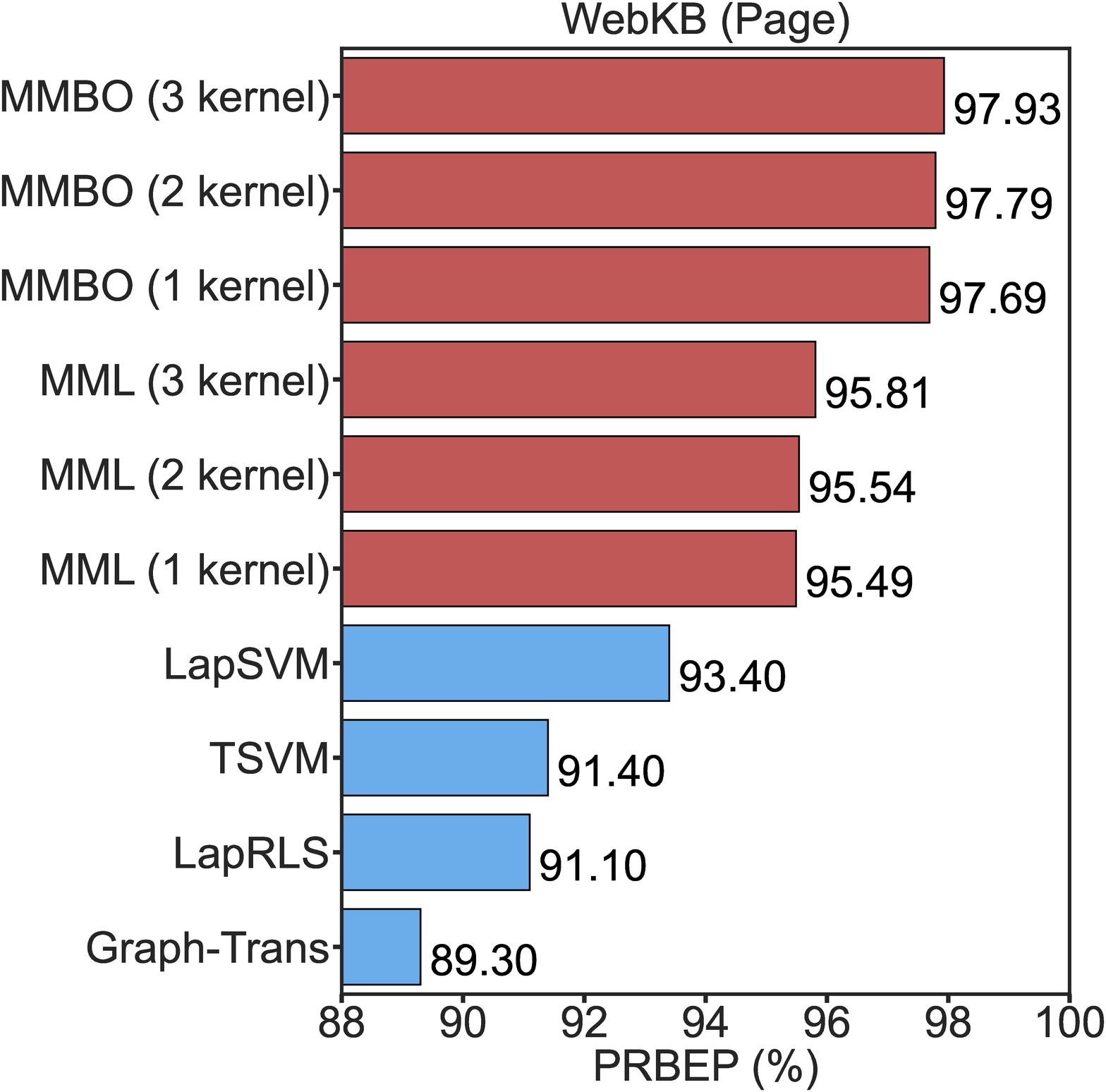}
  \label{fig:sub2}
\end{subfigure}%
\hspace{-2cm}
\begin{subfigure}{.4\textwidth}
  \centering
  \includegraphics[width=.7\linewidth]{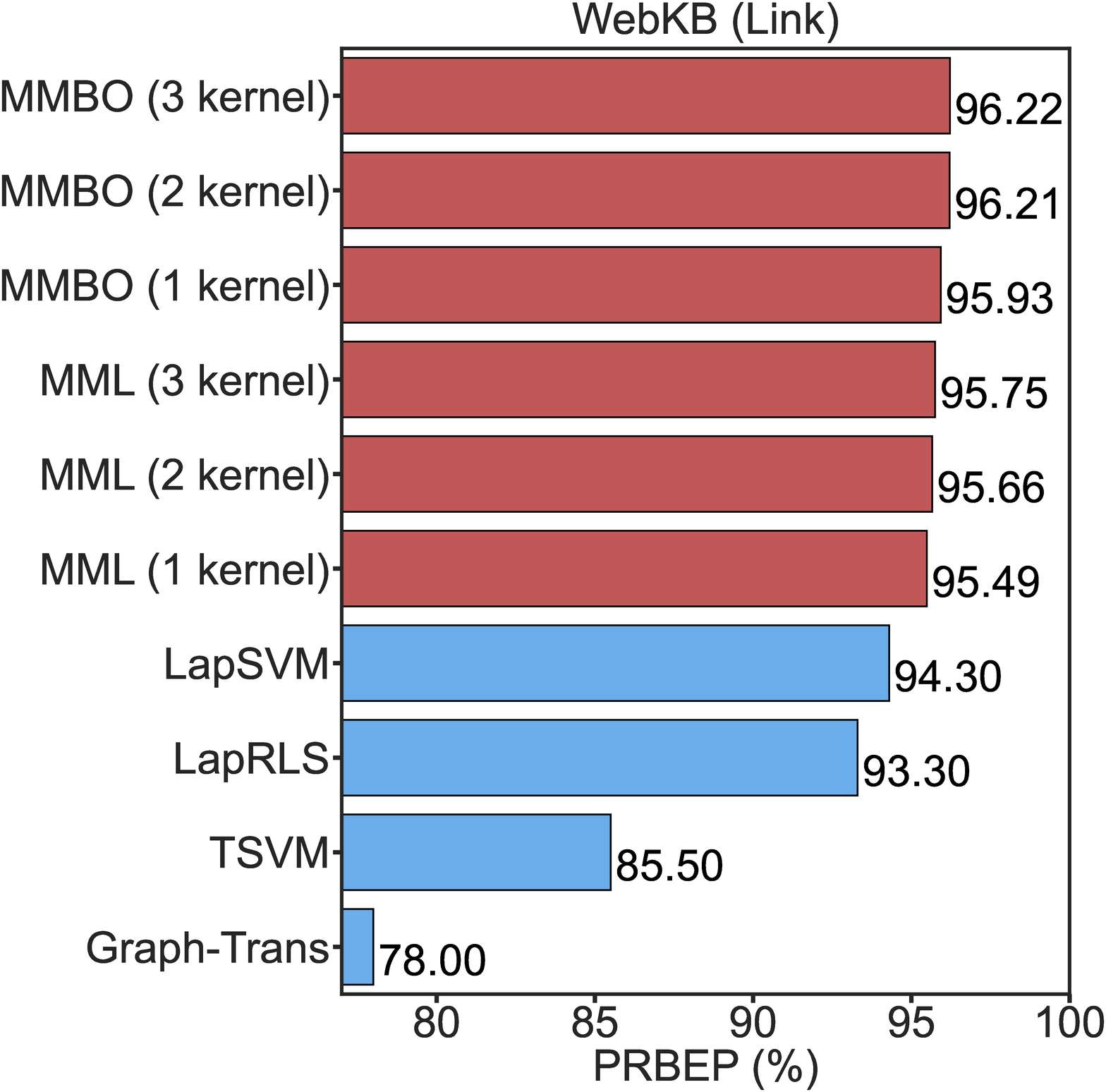}
  \label{fig:sub2}
\end{subfigure}
%

\vspace{0.2cm}
\caption{Comparison of MML and MMBO with other methods on non-biological data. The proposed methods are in red, and other methods are in blue. We note that some of the comparison methods for USPST use more labeled samples than the proposed methods. Please refer to Section \ref{comparison} for more details.}
\label{fig:results}
\end{figure*}

 
 The results for non-biological data sets are shown in Figure \ref{fig:results}. In most experiments with non-biological data, the proposed MMBO method is clearly the most dominant. The other proposed model, the MML method, is the second best model with promising performances.

\subsubsection{Alpha and beta protein classification}

\begin{figure*}
\centerline{\includegraphics[keepaspectratio,width=5.3in]{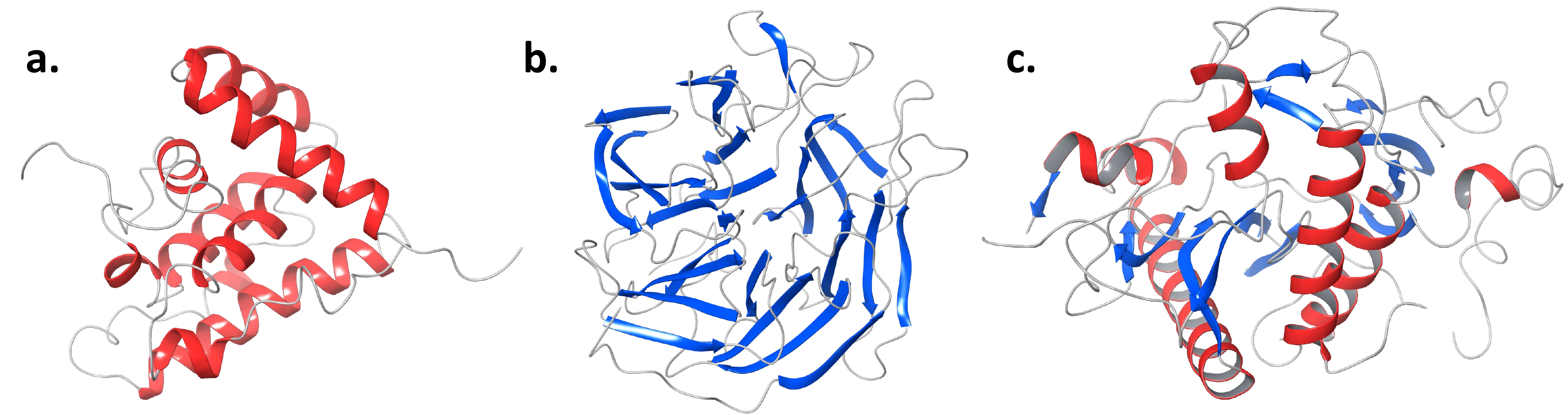}}
\vspace{0.2cm}
\caption{Secondary-structure representations of proteins taken from $\alpha,\beta$-protein data. Here, alpha helix is colored in red, beta sheet is colored in blue. a) Alpha protein (PDBID: 1WIX), b) Beta protein (PDBID: 3O4P), c) Mixed-alpha and beta protein (PDBID: 2CNQ). PDBID stands for protein data bank ID with experimental structures available at https://www.rcsb.org/.}
\label{fig:protein}
\end{figure*}

We also tested the proposed multiscale learning models using biological data, such as data involving protein classification. 
In this data, based on the secondary structure, proteins are typically grouped into three classes, namely alpha helices, beta sheets, and mixed alpha and beta domains. Figure \ref{fig:protein} plots the secondary-structure representations of 3 types of protein structures.
The data, which consists of 900 structures equally distributed into three classes, was collected by Cang et al \cite{ZXCang:2015} and taken from SCOPe (Structural Classification of Proteins-extended), an online database \cite{Fox:2014}. 

Five-fold cross validation is conducted to examine the performance of the proposed models. To preserve the unbiased information, in each fold, the test set consisted of 180 instances with 60 samples from each group. Overall, the protein data sets originally provide the coordinates and atom types for each structure. However, feature generation is needed to translate such information to a vector format suitable for machine learning algorithms. Moreover, for this data, the feature generation has to sustain crucial physical and chemical interactions such as covalent and non-covalent bonds, electrostatic, hydrogen bonds, etc. In the past few years, we have developed numerous mathematical-based feature engineering models including geometric and algebraic graph \cite{DDNguyen:2017d,nguyen2019agl}, differential geometry \cite{nguyen2019dg}, persistent homology \cite{cang2018representability}, and persistent graph \cite{wang2020persistent} for representing 3D molecular information in low dimensional representations.

We employ our geometric graph representation in \cite{DDNguyen:2017d}. In order to represent the physical and chemical properties of a biomolecule, we consider four atom types, namely C$_\alpha$, C, N, and O. In particular, the protein structures are described by vectors of 50 components. Overall, the details of the parameters for the feature generated approach is provided in the Supporting Information. 
 
\begin{figure*}
\centerline{\includegraphics[keepaspectratio,width=3.5in]{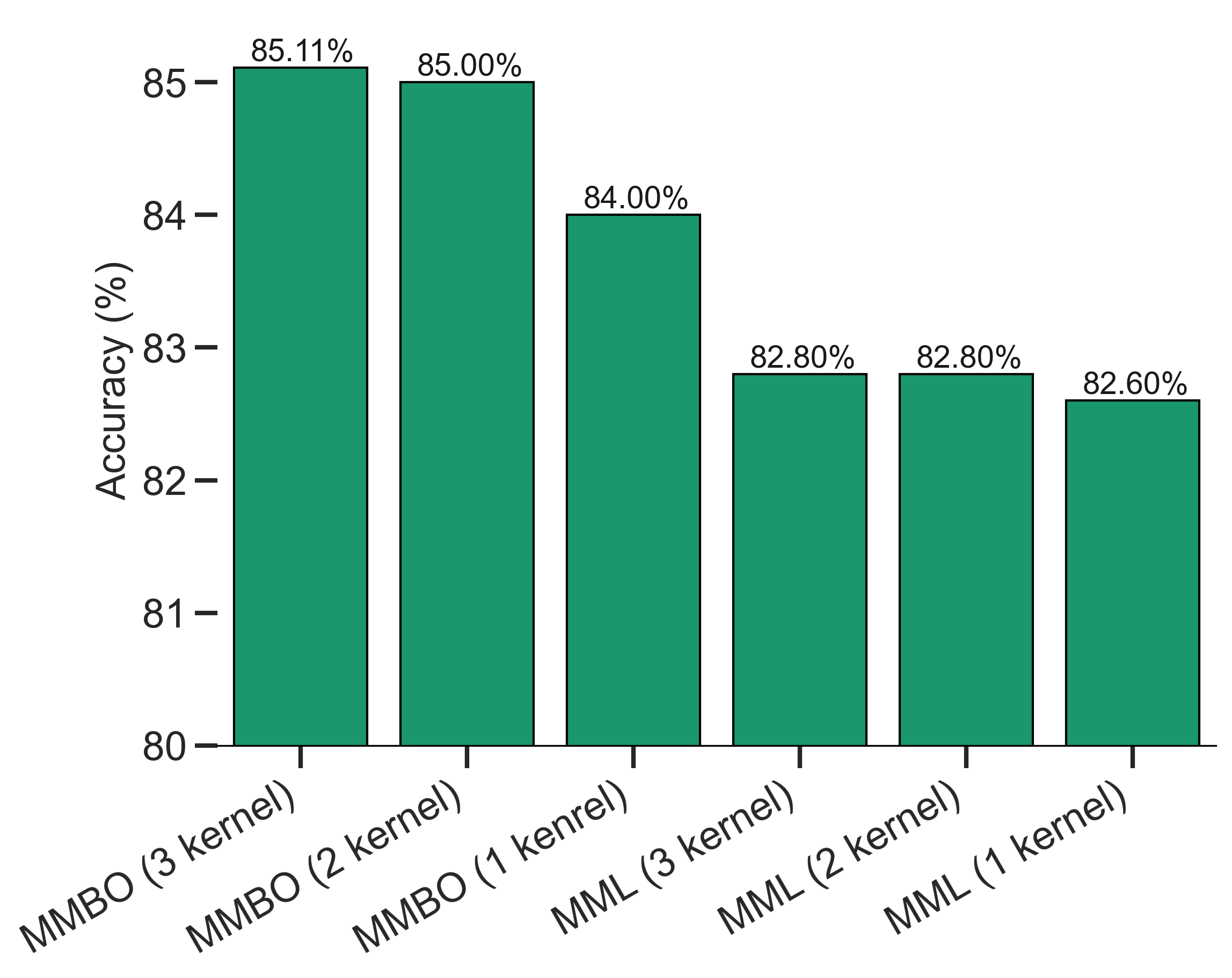}}
\vspace{0.2cm}
\caption{The performances of MMBO and MML models on the protein classification data set.}
\label{fig:protein_results}
\end{figure*}

Both MMBO and MML models perform well. Moreover, similarly to previous experiments, multiscale information strengthens the accuracy of both MML and MMBO approaches. In fact, there is an encouraged improvement from the one kernel model to the two kernel model , i.e., 84\% to 85\% accuracy for the MMBO model. There is also an improvement in the MMBO method results using three kernels, i.e. 85.11\%. For the MML method, there is a slight improvement by using multiple kernels. For this data, the MMBO method outperforms its counterpart, which indicates the versatility of the MMBO algorithm when dealing with a variety of data. All results are presented in Figure \ref{fig:protein_results}.
 
\vspace{0.2cm}

\subsection{Comparison Algorithms}
\label{comparison}

We compare our algorithms to many recent methods, most of which are from 2015 and later.

 For WebKB data, we compare classification accuracy against recent methods such as semi-supervised multi-view deep discriminant representation learning (SMDDRL) \cite{webkb__2}, vertical ensemble co-training (VE-CoT) \cite{webkb__3}, auto-weighted multiple graph learning (AMGL) \cite{webkb__4}, multi-view learning with adaptive neighbors (MLAN) \cite{webkb__5}, deep canonically correlated autoencoder (DCCAE) \cite{webkb__7}, multi-view discriminative neural network (MDNN) \cite{webkb__8}, semi-supervised learning for multiple graphs by gradient flow (MGSC) \cite{webkb__9}, multi-domain classification w/ domain selection (MCS) \cite{webkb__10}, multi-view semi-supervised learning (FMSSL, FMSSL-K) \cite{webkb__11}, and semi-supervised multimodal deep learning framework (SMDLF) \cite{webkb__6}. Our results are obtained using 105 labels, and using the classification accuracy metric. Results for SDMDRL, VE-CoT, AMGL, MLAN, SMDLF, DCCAE and MDNN are from \cite{webkb__2}, the results for MGSC and MCS are from \cite{webkb__9}, and the results for FMSSL and FMSSL-K are from \cite{webkb__11}. All methods use 105 labels.

For USPST, we compare against recent methods such as transductive minimax probability machines (TMPM) \cite{uspst__1}, semi-supervised extreme learning machines (SS-ELM) \cite{uspst__4}, graph embedding-based dimension reduction with extreme learning machines (GDR-ELM) \cite{uspst__6}, extreme learning machine auto-encoder (ELM- AE) \cite{uspst__7}, extreme learning machine auto-encoder with invertible functions (ELM- AEIF) \cite{uspst__8} and extreme learning machines for dimensionality reduction (SR-ELM) \cite{uspst__9}. Our results are obtained using only 50 labels. The results for TMPM (with 50 labels) are from \cite{uspst__1}, the results for GDR-ELM, ELM-AE, ELM-AEIF and SR-ELM (with 150 labels) are from \cite{uspst__6}, and the result for SS-ELM (with 100 labels) are from \cite{uspst__4}.

For G50C, we compare against recent methods such as classtering (CLSST) \cite{g50c__1}, semi-supervised broad learning system (SS- BLS) \cite{g50c__2}, classification from positive and unlabeled data (PNU) \cite{g50c__3}, classification from unlabeled positive and negative data (PUNU) \cite{g50c__3}, semi-supervised extreme learning machines (SS-ELM) \cite{uspst__4}, semi-supervised hierarchical extreme learning machine (SS-HELM) \cite{uspst__2}, safe semi-supervised support vector
machines (S4VM) \cite{g50c__4}, robust and fast transductive support vector machines (RTSVM, RTSVM- LDS) \cite{g50c__5}. Our results are obtained using 50 labels. The result for CLSST is from \cite{g50c__1}, the results for SS-BLS, SS-ELM and SS-HELM are obtained from \cite{g50c__2}, the results for PNU, PUNU and S4VM are obtained from \cite{g50c__3}, and the results for RTSVM and RTSVM-LDS are obtained from \cite{g50c__5}. All comparison methods use 50 labels.

For Mac-Win, we compare against recent methods such as support vector machines with manifold regularization and partially labeling privacy protection (SVM-MR\&PLPP) \cite{macwin__2} and a scalable version (SSVM-MR\&PLPP) \cite{macwin__2}. These results are obtained from \cite{macwin__2}. All comparison methods and the proposed algorithms use 50 labels, the same number of labeled samples as the proposed methods.

We also compare results for all data sets with slightly older methods such as transductive graph methods (Graph-Trans), closely related to \cite{all__2, zhou:bousquet:lal, all__3}, transductive support vector machines (TSVM) \cite{all__1}, support vector machines on a graph-distance derived kernel (Graph-density)  \cite{svm1}, TSVM by gradient descent ($\nabla$TSVM)  \cite{svm1}, low density separation (LDS) \cite{svm1}, Laplacian support vector machines (LapSVM) \cite{sind}  and Laplacian regularized least squares (LapRLS) \cite{sind}. For WebKB, we use the PRBEP metric when comparing against these methods. The results for all older methods, except LapSVM and LapRLF, are obtained from \cite{svm1}, the results for LapSVM and LapRLF are from \cite{sind}. All comparisons with older methods use the same number of labeled samples as the proposed methods: 12 labels for WeBKB and the PRBEP metric, and 50 labels for the rest of the data. \\

 \begin{table*}[!ht]
	\centering
	{
		\caption{The timing of the proposed MMBO method}
		\label{tab:MMBO_timing}
						\vspace{0.05cm}
		\begin{tabular}{lcccc}
			\hline
			Data set & Size of &  Sample & Timing  & Timing  \\
                                      &  data set &    dimension    & (Construction of graph  &   (MMBO procedure) \\
                                      &   &     &  and eigenvectors) &    \\
			\hline
			G50C   & 550&  50    & 0.02 seconds & 0.31 seconds \\
			USPST &   1440 &  1024  & 1.41 seconds  & 1.52 seconds \\
		         Mac-Win &  1946    &   7511  & 9.8 seconds & 1.17 seconds \\
		         WebKB (page) & 1051         &  3000      & 1.04 seconds & 0.60 seconds \\
			WebKB (link) & 1051         &  1840       & 0.67 seconds & 0.60 seconds\\
			WebKB (page+link) & 1051    &  4840       & 1.58 seconds & 0.60 seconds \\        
		        $\alpha,\beta$-protein &   900 & 50  & 0.18 seconds & 1.96 seconds   \\
			\hline 
		\end{tabular}
	}
\end{table*}

          
\begin{table*}[!ht]
\centering
		\caption{The timing of the proposed MML method}
		\label{tab:MML_timing}
						\vspace{0.05cm}
		\begin{tabular}{lcccc}
			\hline
			Data set & Size of &  Sample & Timing  & Timing  \\
                                      &  data set &    dimension    & (Deformed Kernel) &   (Optimization) \\
			\hline
			G50C   & 550&  50    & 0.039 seconds & 0.001 seconds \\
			USPST &   1440 &  1024  & 0.24 seconds  & 0.003 seconds \\
		     Mac-Win &  1946    &   7511  & 4.51 seconds & 0.002 seconds \\
		     WebKB (page) & 1051         &  3000      & 0.21 seconds & 0.02 seconds\\
			WebKB (link) & 1051         &  1840       & 0.15 seconds & 0.01 seconds\\
			WebKB (page+link) & 1051    &  4840       & 0.28 seconds & 0.02 seconds \\        
					$\alpha,\beta$-protein &   900 & 50  & 0.05 seconds & 0.02 seconds   \\
			\hline 
		\end{tabular}
\end{table*}

\subsection{Efficiency}

     The proposed MML and MMBO procedures are very efficient. The timing results are listed for all data sets in Table \ref{tab:MMBO_timing} (for MMBO) and Table \ref{tab:MML_timing} (for MML). 
     
     The timing of the proposed MMBO method is divided into two parts: (1) the timing for the construction of the graph weights and the calculation of the extremal eigenvectors of the multiscale graph Laplacian, and (2) the timing of the MMBO procedure. From Table 2, one can see that the proposed MMBO procedure takes under 2 seconds for all data sets, and the graph construction and computation of the eigenvectors takes little time as well. 
     
     The timing of the proposed MML method consists of two categories: (1) the timing for the construction of the warped kernels, and (2) the timing of the optimizer. One can see from Table \ref{tab:MML_timing} that the procedure of generating the multiscale graph and the warped kernel is the most time-consuming step of the MML algorithm, but it is still under 5 seconds when handling the Mac-Win data set having 1946 samples with a feature dimension of 7511. For other data sets, the MML method takes under 0.3 seconds to formulate the multiscale graph and the warped kernel. Due to the simplified version of the optimizer of the MML method, one can directly use the standard solver of SVM for the MML algorithm. This procedure is extremely fast and needs no more than 0.03 seconds to complete the task for all experiments. The computations were performed on a personal laptop 2.4 GHz 8-Core Intel Core i9.

\section{Conclusion}
\label{conc}

This work presents several methods for machine learning tasks and for dealing with some of the challenges of machine learning, such as data with limited samples, smaller data sets, and diverse data, usually associated with small data sets or data related to areas of study where the size of the data sets is constrained by the complexity and/or high cost of experiments. In particular, we integrate graph-based techniques, multiscale structure, adapted and modified optimization procedures and semi-supervised frameworks to derive two multiscale Laplacian learning (MLL) approaches for machine learning tasks, such as data classification. 

The first approach introduces a multiscale kernel representation to a manifold learning technique and is called the multikernel manifold learning (MML) algorithm . 

The second approach combines multiscale analysis with an interesting adaptation and modification of the famous classical Merriman-Bence-Osher (MBO) scheme, originally intended to approximate motion by mean curvature, and is called the multiscale MBO (MMBO) algorithm. 

The performance of the proposed MLL approaches  is favorably compared to existing recent and related approaches through experiments on a variety of data sets. The two new MLL methods form powerful techniques for dealing with some of the most important challenges and tasks in machine learning and data science.

\section*{Supporting Information}
We present the optimal hyperparameters of the proposed MMBO and MML methods for all experiments conducted in this work in Online Resource: Supporting Information.

\section*{Availability}
The source code for the proposed MMMBO and MML methods is available at Github: \url{https://github.com/ddnguyenmath/Multiscale-Laplacian-Learning}.

\section*{Conflict of interest}
The authors declare that they have no conflict of interest.  \\


\end{document}